\documentclass{article}
\usepackage{arxiv}
\usepackage[utf8]{inputenc} % allow utf-8 input
\usepackage[T1]{fontenc}    % use 8-bit T1 fonts
\usepackage{url}  
\usepackage{hyperref}       % hyperlinks
\hypersetup{
    colorlinks=true,       % false: boxed links; true:
    linkcolor=blue,          % color of internal links
    citecolor=blue,        % color of links to 
    urlcolor=blue           % color of external links
}
\usepackage{booktabs}       % professional-quality tables math symbols
\usepackage{nicefrac}       % compact symbols for 1/2, etc.
\usepackage{microtype}      % microtypography
\usepackage{lipsum}		% Can be removed after putting your text content
\usepackage{graphicx}
\usepackage{doi}
\usepackage{amsmath}
\usepackage{subfig}
\usepackage[affil-it]{authblk}
\usepackage[english]{babel}
\usepackage{blindtext}
\usepackage{amssymb}
\usepackage{cleveref}
\usepackage{booktabs}
\usepackage{framed,multirow}
\usepackage{hyperref}
\usepackage{url}            % simple URL typesetting
\usepackage{booktabs}% professional-quality tables
\usepackage[mathlines]{lineno}
\usepackage{amsfonts}       % blackboard math symbols
\usepackage{latexsym}
\usepackage{amsmath}
\usepackage{amssymb}
\usepackage{amsfonts}

\usepackage{cleveref}
\usepackage{nicefrac}       % compact symbols for 1/2, etc.
\usepackage{microtype}      % microtypography
\usepackage{lipsum}
\usepackage{graphicx}
\usepackage{adjustbox}
\usepackage{graphicx}
\usepackage[table,xcdraw]{xcolor}
\definecolor{newcolor}{rgb}{.8,.349,.1}
\usepackage[linesnumbered,ruled,vlined]{algorithm2e}

\usepackage{multirow}
\usepackage{diagbox}
\usepackage{subfig}
\usepackage{tikz}
\usetikzlibrary{external}

\usepackage{pgfplots}
\pgfplotsset{compat=1.7}
\usetikzlibrary{positioning}
\usepackage{etoolbox} 
\usepackage[numbers]{natbib}

\newcommand*\linenomathpatch[1]{%
  \cspreto{#1}{\linenomath}%
  \cspreto{#1*}{\linenomath}%
  \csappto{end#1}{\endlinenomath}%
  \csappto{end#1*}{\endlinenomath}%
}
\newcommand*\linenomathpatchAMS[1]{%
  \cspreto{#1}{\linenomathAMS}%
  \cspreto{#1*}{\linenomathAMS}%
  \csappto{end#1}{\endlinenomath}%
  \csappto{end#1*}{\endlinenomath}%
}
\expandafter\ifx\linenomath\linenomathWithnumbers
  \let\linenomathAMS\linenomathWithnumbers
  \patchcmd\linenomathAMS{\advance\postdisplaypenalty\linenopenalty}{}{}{}
\else
  \let\linenomathAMS\linenomathNonumbers
\fi

\linenomathpatch{equation}
\linenomathpatchAMS{gather}
\linenomathpatchAMS{multline}
\linenomathpatchAMS{align}
\linenomathpatchAMS{alignat}
\linenomathpatchAMS{flalign}

\SetCommentSty{mycommfont}

\crefformat{section}{\S#2#1#3}
\crefformat{subsection}{\S#2#1#3}
\crefformat{subsubsection}{\S#2#1#3}
\crefrangeformat{section}{\S\S#3#1#4 to~#5#2#6}
\crefmultiformat{section}{\S\S#2#1#3}{ and~#2#1#3}{, #2#1#3}{ and~#2#1#3}
\DeclareMathOperator{\sech}{sech}

\date{}
\title{An adaptive augmented Lagrangian method for training physics and equality constrained artificial neural networks}

\author{\href{https://orcid.org/0000-0002-1095-0881}{\includegraphics[scale=0.08]{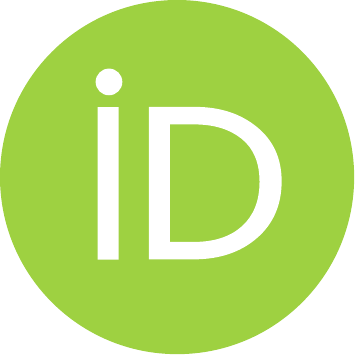}\hspace{1mm}Shamsulhaq Basir}\thanks{~shb105@pitt.edu (Shamsulhaq Basir)}}

\author{\href{https://orcid.org/0000-0003-1967-7583}{\includegraphics[scale=0.08]{orcid.pdf}\hspace{1mm}Inanc Senocak\thanks{corresponding author:~senocak@pitt.edu (Inanc Senocak)}}}

\affil{Department of Mechanical Engineering and Materials Science, University of Pittsburgh, \\ 3700 O'Hara St., Pittsburgh, PA 15261, USA}
\renewcommand{\shorttitle}{PECANNs with Adaptive ALM}

\begin{document}
\maketitle
%% Abstract %%
\begin{abstract}
Physics and equality constrained artificial neural networks (PECANN) are grounded in methods of constrained optimization to properly constrain the solution of partial differential equations (PDEs) with their boundary and initial conditions and any high-fidelity data that may be available. To this end, adoption of the augmented Lagrangian method within the PECANN framework is paramount for learning the solution of PDEs without manually balancing the individual loss terms in the objective function used for determining the parameters of the neural network. Generally speaking, ALM combines the merits of the penalty and Lagrange multiplier methods while avoiding the ill conditioning and convergence issues associated singly with these methods . In the present work, we apply our PECANN framework to solve forward and inverse problems that have an expanded and diverse set of constraints. We show that ALM with its conventional formulation to update its penalty parameter and Lagrange multipliers stalls for such challenging problems. To address this issue, we propose an adaptive ALM in which each constraint is assigned a unique penalty parameter that evolve adaptively according to a rule inspired by the adaptive subgradient method. Additionally, we revise our PECANN formulation for improved computational efficiency and savings which allows for mini-batch training. We demonstrate the efficacy of our proposed approach by solving several forward and PDE-constrained inverse problems with noisy data, including simulation of incompressible fluid flows with a primitive-variables formulation of the Navier-Stokes equations up to a Reynolds number of 1000.
\end{abstract}

%% Keywords %%


%%%% Introduction %%% 
\section{Introduction}\label{sec:intro}
Partial differential equations (PDEs) are used to describe a wide range of physical phenomena, including sound propagation, heat and mass transfer, fluid flow, and elasticity. The most common numerical methods for solving PDE problems involve domain discretization, such as finite difference, finite volume, finite element, and spectral element methods. However, the accuracy of the solution heavily depends on the quality of the mesh used for domain discretization. Additionally, mesh generation can be tedious and time-consuming, particularly for complex geometries or problems with moving boundaries. While numerical methods are efficient for solving forward problems, they are not well-suited for solving inverse problems, particularly data-driven modeling. In this regard, neural networks can be viewed as a promising alternative meshless approach to solving PDEs. 

The use of neural networks for solving PDEs can be traced back the early 1990s \cite{dissanayake1994neural,van1995neural,Monterola1998lagrange,lagaris1998artificial}. Recently, there has been a resurgence of interest in the use of neural networks to solve PDEs \cite{E2018,Han2018,sirignano2018dgm, Zhu2019}, particularly after the introduction of the term physics-informed neural networks (PINNs) \cite{raissi2019pinn}. In PINNs, the governing equations of a physical phenomenon of interest are embedded in an objective function along with any data to learn the solution to those governing equations. Although the performance of PINNs can be influenced by several factors such as the choice of activation function \cite{jagtap2020adaptive}, sampling strategy \cite{wu2023comprehensive}, and architecture \cite{wang2021understanding,PECANN_2022}, the formulation of the objective function that is used to determine the neural network parameters is paramount for satisfactory predictions. 

In the baseline PINN formulation \cite{raissi2019pinn},  several loss terms with different physical scales are aggregated into a single composite objective function with tunable weights, which is a rudimentary approach to a multi-objective optimization problem. Since neural networks are trained using gradient descent type optimizers, the model parameters can be influenced by the larger gradient of the loss function, irrespective of the physical scale. This can lead to unstable training, as the optimizer may prioritize one objective over another, sacrificing either the PDE loss or the boundary loss. Therefore, adjusting the interplay between the objective terms requires manual hyperparameter tuning, which can be a time-consuming and challenging task. Additionally, the absence of validation data or prior knowledge of the solution to the PDE for the purpose of tuning can render the baseline PINN approach impractical for the solution of PDEs \cite{basir2022critical}. 

%Unfortunately, it is not uncommon for researchers in the scientific community to overlook or underestimate the issue of hyperparameter selection by resorting to arbitrary choices based on prior knowledge of the solution or an attempt to fit it. It is important to emphasize that having prior knowledge of the solution renders the use of any solver unnecessary. Therefore, while PINNs show great potential for solving PDEs, their practicality depends on addressing these technical issues.
 Dynamic determination of the weights in the composite objective function of the baseline PINN approach has attracted the attention of several researchers.  \citet{wang2021understanding} proposed an empirical method that has several limitations that we discussed in a prior work \cite{PECANN_2022}. \citet{van2022optimally} proposed an empirical method by considering a bi-objective loss function for the solution of linear PDEs. \citet{liu2021dual} proposed a dual dimer method to adjust the interplay between the loss terms by searching for a saddle point. 
 \citet{wang_parisi_tangent_kernel} studied PINNs using the Neural Tangent Kernel (NTK). NTK provides insights into the dynamics of fully-connected neural networks with infinite width during training via gradient descent. The authors proposed using the eigenvalues of the NTK to adapt the interplay between the objective terms. \citet{mcclenny2020self} proposed self-adaptive PINNs (SA-PINN) with a minimax formulation to adjust the interplay between the loss terms. Lagrange multipliers are sent through an empirical mask function such as sigmoid, which makes the dual unconstrained optimization formulation not equivalent to the constrained optimization problem. Because of that, equality constraints are not strictly enforced. The SA-PINN method produced results that are comparable or better than the NTK method \cite{wang2021understanding}. In the present work, we compare our proposed method with both of these approaches for the solution of the wave equation. %\citet{mojgani2022lagrangian} proposed a change of frame-of-reference from Eulerian to Lagrangian for the solution of convection-diffusion equations. However, the core difficulty in training artificial neural networks for scientific problems lies in properly formulating an optimization problem that strictly and adaptively enforces physics constraint without the need for manual tuning.
 
By and larger the aforementioned works adopt an unconstrained optimization approach in the first place to formulate the objective function used to determine neural network parameters. In physics and equality constrained artificial neural networks (PECANNs) \cite{PECANN_2022}, we pursued a constrained optimization method to formulate the objective function. Specifically, we used the augmented Lagrangian method (ALM) \cite{hestenes1969multiplier, powell1969method} to formally cast the constrained optimization problem into an unconstrained one in which PDE domain loss is constrained by boundary and initial conditions, and with any high-fidelity data that may be available. It is worth noting that ALM combines the merits of the penalty method and the Lagrange multiplier method. It balances feasibility and optimality by updating a penalty parameter to control the influence of constraint violations \cite{nocedal2006numerical}. 

In what follows, we show that the conventional ALM with a single penalty parameter used in the PECANN model \citep{PECANN_2022} struggles when applied to problems with multiple constraints of varying characteristics. To overcome this limitation, we propose an adaptive augmented Lagrangian method, which introduces multiple penalty parameters and independently updates them based on the characteristics of each constraint. Additionally, we propose a computationally efficient formulation of the objective function to handle a large number of constraints to enable mini-batch training while maintaining a small memory footprint during training. We solve several forward and inverse PDE problems, including the solution of incompressible fluid flow equations with a primitive-variables formulation up to a Reynolds number of 1000, to demonstrate the efficacy of our PECANN model with an adaptive augmented Lagrangian method. The codes used to produce the results in this paper are publicly available at \url{https://github.com/HiPerSimLab/PECANN}

%%%% Technical Background %%% 
\section{Technical formulation}\label{sec:tech_back}
Let us consider a general constrained optimization problem with equality constraints
\begin{equation}
\begin{aligned}
    \min_{\theta} \mathcal{J}(\theta), ~\quad \text{such that } ~\quad \mathcal{C}_i(\theta) =0, \quad \forall i \in \mathcal{E}, \label{eq:constrained_problem}
\end{aligned}
\end{equation}
where the objective function $\mathcal{J}$ and the constraint functions $\mathcal{C}_i$ are all smooth, real valued functions on a subset of $R^n$ and $\mathcal{E}$ is a finite set of equality constraints. We can write an equivalent \emph{minimax} unconstrained dual problem using the augmented Lagrangian method as follows 
\begin{align}
 \min_\theta \max_\lambda \mathcal{L}(\theta;\lambda,\mu) = \mathcal{J}(\theta) + \sum_{i \in \mathcal{E}} \lambda_i \mathcal{C}_i+ \frac{\mu}{2} \sum_{i \in \mathcal{E}} \mathcal{C}^2_i (\theta).
\label{eq:max_min_multiplier}
\end{align} 
We can swap the order of the minimum and the maximum by using the following \textit{minimax} inequality concept or weak duality
\begin{align}
   \max_\lambda \min_\theta {L}(\theta;\lambda)  \le \min_\theta \max_\lambda {L}(\theta;\lambda).
\end{align}

The minimization can be performed for a sequence of multipliers generated by
\begin{subequations}
\begin{align}
    \lambda_i &\leftarrow \lambda_i + \mu ~ \mathcal{C}_i(\theta),~\quad \forall i \in \mathcal{E}.
    \label{eq:dual_ascent_aug}
\end{align}
\end{subequations}
We should note that $\mu$ can be viewed as a global learning rate for the Lagrange multipliers. ALM combines the merits of the penalty method and the Lagrange multiplier method by updating the penalty parameter in such a way that it balances the trade-off between feasibility and optimality, and ensures convergence. Traditionally, in ALM, textbook descriptions typically rely on a single penalty parameter to address all constraints as in \eqref{eq:dual_ascent_aug}, and updating the penalty parameter is often done through empirical strategies. However, as we show in a later section, these update strategies become ineffective when there are multiple constraints with different characteristics. Next, we consider two existing strategies to update the penalty parameter in ALM. 

First, in Algorithm \ref{alg:classic_training_algorithm}, we monotonically increase the penalty parameter $\mu$ at each training iteration until a maximum safeguarding value (i.e., $\mu_{\max}$) is reached. Safeguarding the penalty parameter is a common strategy and prevents it from reaching excessively large values that may lead to numerical instability or overflow. To establish clear distinction from other strategies and facilitate comparison, we will refer to Algorithm \ref{alg:classic_training_algorithm} as ALM with monotonically increasing penalty update (MPU).
%training algoritm
\IncMargin{1em}
\begin{algorithm}[!h]
\SetAlgoLined
% setting keywords 
\SetKw{KwInput}{Input:}
\SetKw{KwOutput}{Output:}
\SetKw{KwDefaults}{Defaults:}
% algorithm 
\KwInput{$\theta^0,\mu_{\max},\beta$}\\
$\lambda_{i}^0 = 1 ~ \forall i \in \mathcal{E}$ \hspace{1em}
\tcc{Initializing Lagrange multipliers}
$\mu^0 = 1 $ \hspace{4em}
\tcc{Initializing the penalty parameter}
\BlankLine
\For{$t = 1  ~ \KwTo ...$}{
    $\theta^t \leftarrow \underset{\theta}{\mathrm{argmin}}~ \mathcal{L}(\theta^{t-1};\lambda^{t-1},\mu^{t-1})$\hspace{1em}
    \tcc{primal update}
    $\lambda_i^t \xleftarrow{} \lambda_i^{t-1} + \mu^{t-1} \mathcal{C}_i(\theta^t)$ \hspace{4em}
    \tcc{update Lagrange multipliers}
        $\mu^t = \min( \beta~\mu^{t-1}, \mu_{\max})$ \hspace{4em}
        \tcc{penalty parameter update}
}
\KwOutput{$\theta^t$}\\
\KwDefaults{$\beta = 2, ~\mu_{\max} = 1 \times 10^4$}\\
\caption{ALM with monotonically increasing penalty update (MPU) \cite{martins2021engineering}}\label{alg:classic_training_algorithm}
\end{algorithm}
%%%

In the context of training the neural networks, inputs to Algorithm \ref{alg:classic_training_algorithm} are parameters for the neural network model $\theta^0$, a maximum safeguarding penalty parameter $\mu_{\max}$, and a multiplicative factor $\beta$ for increasing the penalty parameter $\mu$ at each iteration.  We should note that in Algorithm \ref{alg:classic_training_algorithm}, both the Lagrange multipliers and the penalty parameter are updated at each iteration \cite{martins2021engineering}. A similar approach without the maximum safeguarding penalty parameter has been used in \cite{lu2021}. To prevent divergence, the maximum penalty parameter was set to $10^4$. However, updating the penalty parameter at each epoch may aggressively increase the penalty parameter that could lead to divergence as we will demonstrate in a later section. Additionally, finding a suitable maximum penalty parameter can be challenging. 

Another strategy to update the penalty parameter $\mu$ is to update it only when the constraints have not decreased sufficiently at the current iteration \cite{bierlaire2015optimization,wright1999numerical}. In Algorithm 
\ref{alg:standard_training_algorithm}, we present an augmented Lagrangian method in which the penalty parameter is updated conditionally \cite{wright1999numerical}. A similar strategy has been adopted in the work of \citet{dener2020training} to train an encoder-decoder neural network for approximating the Fokker-Planck-Landau collision operator. We will refer to Algorithm \ref{alg:standard_training_algorithm} as ALM with conditional penalty update (MPU).
%training algoritm
\IncMargin{1em}
\begin{algorithm}[!h]
\SetAlgoLined
% setting keywords 
\SetKw{KwInput}{Input:}
\SetKw{KwOutput}{Output:}
\SetKw{KwDefaults}{Defaults:}
% algorithm 
\KwInput{$\theta^0,\mu_{\max},\beta$}\\
$\lambda_{i}^0 = 0 ~ \forall i \in \mathcal{E}$ \hspace{1em}
\tcc{Initializing Lagrange multipliers}
$\mu^0 = 1 $ \hspace{4em}
\tcc{Initializing the penalty parameter}
\BlankLine
\For{$epoch = 1  ~ \KwTo ...$}{
    $\theta^t \leftarrow \underset{\theta}{\mathrm{argmin}}~ \mathcal{L}(\theta^{t-1};\lambda^{t-1},\mu^{t-1})$\hspace{4em}
    \tcc{primal update}
    \uIf{$(\|\mathcal{C}(\theta^t)\|_2 < ~ \eta^{t-1}$}{
        $\lambda_i^t \xleftarrow{} \lambda_i^{t-1} + \mu^{t-1} \mathcal{C}_i(\theta^t)$ \hspace{3em}
        \tcc{update Lagrange multipliers}
        $\mu^t = \mu^{t-1}$ \hspace{8em}
        \tcc{the penalty parameter unchanged}
        }
    \Else{
        $\mu^t = \min( \beta~\mu^{t-1}, \mu_{\max})$ \hspace{0em}
        \tcc{penalty parameter update}
        $\lambda_i^t \xleftarrow{} \lambda_i^{t-1}$ \hspace{6em}
        \tcc{Lagrange multipliers unchanged}
      }
      $\eta^t = \|\mathcal{C}(\theta^t)\|_2$
}
\KwOutput{$\theta^t$}\\
\KwDefaults{$\beta = 2, ~\mu_{\max} = 1 \times 10^4$}\\ \caption{ALM with conditional penalty update (CPU) \cite{wright1999numerical}}
 \label{alg:standard_training_algorithm}
\end{algorithm}
Our inputs to Algorithm \ref{alg:standard_training_algorithm} are $\eta$ which is a placeholder for the previous value of constraints, $\mu_{\max}$ is a safeguarding penalty parameter, $\beta$ is a multiplicative weight for increasing penalty parameter. Similar to the previous algorithm, the maximum penalty parameter is set to $10^4$ to prevent numerical overflow and divergence. It should be noted that finding a suitable maximum penalty parameter can be challenging.

In our recent investigations, we have discovered that the rate of update for Lagrange multipliers can be a critical factor when dealing with problems that have different types of constraints (such as flux, data, or PDE constraints). Because the penalty parameter behaves like a learning rate for the Lagrange multiplier, employing the same update rate or penalty parameter for all constraints can lead to issues of instability during training. In some cases, the parameter may be too large or too small for certain constraints, which can adversely affect the optimization process. As such, we propose a new approach to address this issue. Our method involves assigning a unique penalty parameter for each constraint type with its own update rate. By tailoring the penalty parameter to each specific constraint type, we can effectively manage the optimization process and ensure greater stability during training. We will show in the results section that we are able to learn the solution of challenging PDEs when other strategies fail or perform poorly.

\subsection{Adaptive augmented Lagrangian method}
In this section, we propose an augmented Lagrangian method with a novel adaptive update strategy for multiple penalty parameters to handle diverse constraints in the optimization problem. We note that it is common to use a single penalty parameter that is monotonically increased in most implementations of the ALM. However, that strategy becomes insufficient to tackle challenging problems. As we discuss above, we need a unique, adaptive penalty parameter or learning rate for each Lagrange multiplier associated with a constraint. This tailored approach ensures that the penalty parameter conforms to the characteristics of individual Lagrange multipliers, enabling effective handling of diverse constraints. We formulate our unconstrained optimization problem for problem \eqref{eq:constrained_problem} as follows:
\begin{align}
    \max_{\lambda} \min_{\theta} \mathcal{L}(\theta,\lambda;\mu) =  \mathcal{J}(\theta) + \sum_{i \in \mathcal{E}} \lambda_i \mathcal{C}_i(\theta)  + \frac{1}{2} \sum_{i \in \mathcal{E}} \mu_i  \mathcal{C}^2_i(\theta),
\end{align}
where $\lambda_{i}$ is a vector of Lagrange multipliers and $\mu_{i}$ is a vector of penalty parameters. The minimization can be performed using a variant of gradient descent type optimizer for a sequence of Lagrange multipliers generated by 
\begin{align}
    \lambda_i &\xleftarrow{} \lambda_i +  \mu_i \mathcal{C}_i(\theta), && \forall i \in \mathcal{E},
    \label{eq:adpative_multi_penalty_dual_update}
\end{align}
during training. Upon examining the dual update of the augmented Lagrangian method as shown in Eq.~\eqref{eq:adpative_multi_penalty_dual_update}, we observe that it involves a gradient ascent step with learning rates denoted by $\mu_i$ for each Lagrange multiplier $\lambda_i$. Hence, an suitable approach is to adopt the strategy in RMSprop algorithm by G. Hinton \cite{RMSprop_hinton} in finding an independent effective learning rate or penalty parameter for each Lagrange multiplier. The main reason behind our choice is that the update strategy in RMSprop is consistent with the dual update of ALM as in Eq.~\eqref{eq:adpative_multi_penalty_dual_update}. Hence, we divide our global learning rates by the weighted moving average of the squared gradients of our Lagrange multipliers as follows
\begin{align}
    \bar{v}_i & \xleftarrow{} \alpha \bar{v}_i + (1 - \alpha) 
    \mathcal{C}_i(\theta)^2, &&\forall i \in \mathcal{E},\\
    \mu_i & \xleftarrow{} \frac{\gamma}{\sqrt{\bar{v}_i} + \epsilon}, &&\forall i \in \mathcal{E},\\
    \lambda_i &\xleftarrow{} \lambda_i +  \mu_i \mathcal{C}_i(\theta), &&\forall i \in \mathcal{E},
    \label{eq:adaptive_dual_update}
\end{align}
where $\bar{v}_i$ are the weighted moving average of the squared gradient of our Lagrange multipliers, $\gamma$ is a scheduled global learning rate, $\epsilon$ is a term added to the denominator to avoid division by zero for numerical stability, and $\alpha$ is a smoothing constant. In Algorithm \ref{alg:adaptive_training_algorithm}, presents our training procedure. 
% training algoritm
\IncMargin{1em}
\begin{algorithm}[!h]
\SetAlgoLined
% setting keywords 
\SetKw{KwInput}{Input:}
\SetKw{KwOutput}{Output:}
\SetKw{KwDefaults}{Defaults:}
% algorithm 
\KwDefaults{$\gamma = 1\times 10^{-2}, ~\alpha = 0.99,~ \epsilon = 1\times 10^{-8}$}\\

\KwInput{$\theta^0$}\\
$\lambda_{i}^0 = 1 \quad \forall i \in \mathcal{E}$ \hspace{6em}
\tcc{Initializing Lagrange multipliers}
$\mu_{i}^0 = 1  \quad \forall i \in \mathcal{E}$ \hspace{5em}
\tcc{Initializing penalty parameters}
$\bar{v}_{i}^0 = 0 \quad \forall i \in \mathcal{E}$ \hspace{6em}
\tcc{initializing averaged square-gradients}
\BlankLine
\For{$t = 1  ~ \KwTo ...$}{
    $\theta^t \leftarrow \underset{\theta}{\mathrm{argmin}}~ \mathcal{L}(\theta^{t-1};\lambda^{t-1},\mu^{t-1})$\hspace{6em}
    \tcc{primal update}
    $\bar{v}_i^t  \xleftarrow{} \alpha ~\bar{v}_i^{t-1} + (1 - \alpha)~
    \mathcal{C}_i(\theta^t)^2, \quad \forall i \in \mathcal{E}$\hspace{3em}
    \tcc{square-gradient update}
    $\mu_i^{t}  \xleftarrow{} \frac{\gamma}{\sqrt{\bar{v}_i^t} + \epsilon}, \quad \forall i \in \mathcal{E}$\hspace{10.5em}
   \tcc{penalty update}
    $\lambda_i^t \xleftarrow{} \lambda_i^{t-1} +  \mu_i^{t} ~ \mathcal{C}_i(\theta^{t}), \quad \forall i \in \mathcal{E}$ \hspace{6em}
    \tcc{dual update}
    }
\KwOutput{$\theta^t$}\\
 \caption{Augmented Lagrangian method with adaptive penalty updates (APU)}
 \label{alg:adaptive_training_algorithm}
\end{algorithm}
The input to the algorithm is an initialized set of parameters for the network $\theta^{0}$, a global learning rate $\gamma$, a smoothing constant $\alpha$, collocation points, noisy measurement data and high fidelity data to calculate our augmented Lagrangian. In Algorithm \ref{alg:adaptive_training_algorithm}, the Lagrange multiplier vector is initialized to $1.0$ with their respective averaged squared-gradients initialized to zero. In summary, our main approach involves reducing the penalty parameters for Lagrange multiplier updates with large and rapidly changing gradients, while increasing the penalty parameters for Lagrange multiplier updates with small and slowly changing gradients. Next, we delve into the subtle differences in learning the solution of forward and inverse problems and illustrate how they can be formulated as constrained optimization problems, which can then be cast as an unconstrained optimization problem using the augmented Lagrangian method. 

%% proposed formulation
%\section{Proposed Formulation: Forward/Inverse PDE Problems}
\subsection{Constrained optimization formulation for solving forward and inverse problems}
In this section, we explain our proposed formulation for solving a generic PDE problem with boundary and initial conditions. for demonstration purposes. We will also compare our new formulation to our previous approach and highlight the advantages and improvements of our new approach. We first formulate a constraint optimization problem by minimizing the loss on the governing PDE and constraining the individual points on the boundary and initial conditions assuming these conditions are noise-free and can be defined as equality constraints. Consider a scalar function \(u(\boldsymbol{x},t): \mathbb{R}^{d+1} \rightarrow \mathbb{R}\) on the domain \(\Omega \subset \mathbb{R}^d\) with its boundary \(\partial \Omega\) satisfying the following partial differential equation 
\begin{align}
    \mathcal{F}(\boldsymbol{x},t;\frac{\partial u}{\partial t}, \frac{\partial^2 u}{\partial t^2},\cdots,\frac{\partial u}{\partial \boldsymbol{x}}, \frac{\partial^2 u}{\partial \boldsymbol{x}^2},\cdots,\boldsymbol{\nu}) &= 0,\quad\forall (\boldsymbol{x},t) \in \mathcal{U},\label{eq:PDE}\\
    \mathcal{B}(\boldsymbol{x},t,g;u, \frac{\partial u}{\partial \boldsymbol{x}},\cdots) &= 0, \quad\forall (\boldsymbol{x},t) \in \partial \mathcal{U},\label{eq:BC}\\
    \mathcal{I}(\boldsymbol{x},t,h;u,\frac{\partial u}{\partial t},\cdots) &= 0, \quad \forall (\boldsymbol{x},t) \in \Gamma, \label{eq:IC}
\end{align}
where $\mathcal{F}$ is the residual form of the PDE containing differential operators, $\boldsymbol{\nu}$ is a vector PDE parameters, $\mathcal{B}$ is the residual form of the boundary condition containing a source function $g(\boldsymbol{x},t)$ and $\mathcal{I}$ is the residual form of the initial condition containing a source function $h(\boldsymbol{x},t)$. $\mathcal{U} = \{(\boldsymbol{x},t) ~|~\boldsymbol{x} \in \Omega, t = [0,T] \}$, $\partial \mathcal{U} = \{(\boldsymbol{x},t) ~|~\boldsymbol{x} \in \partial \Omega, t = [0,T]\}$ and $\Gamma = \{(\boldsymbol{x},t) ~|~\boldsymbol{x} \in \partial \Omega, t = 0\}$.
% exisiting formulation
\subsubsection{Formulation for problems with point-wise constraints}
\label{sec:constrained_optimization_formulation_pointwise_constraint}
In this section, we present a constrained optimization formulation, as represented in equation \eqref{eq:constrained_problem}, within the context of solving partial differential equations (PDEs). Considering Eq.\eqref{eq:PDE} with its boundary condition \eqref{eq:BC} and initial condition \eqref{eq:IC}, we previously formulated the following constrained optimization problem \cite{PECANN_2022}: 
\begin{align}
    \min_{\theta} \sum_{i=1}^{N_{\mathcal{F}}}\|\mathcal{F}(\boldsymbol{x}^{(i)},t^{(i)};\boldsymbol{\nu},\theta)\|_2^2,
    \label{eq:objective}
\end{align}
subject to 
\begin{align}
    \phi(\mathcal{B}(\boldsymbol{x}^{(i)},t^{(i)},g^{(i)};\theta)) &= 0,~ \forall (\boldsymbol{x}^{(i)},t^{(i)},g^{(i)}) \in \mathcal{\partial U},~ i = 1,\cdots, N_{\mathcal{B}}
    \label{eq:BCconstraints}\\
    \phi(\mathcal{I}(\boldsymbol{x}^{(i)},t^{(i)},h^{(i)};\theta)) &= 0,~ \forall (\boldsymbol{x}^{(i)},t^{(i)},h^{(i)}) \in \Gamma,~ i = 1,\cdots, N_{\mathcal{I}},
    \label{eq:ICconstraints}
\end{align}
where $N_{\mathcal{F}}$, $N_{\mathcal{B}}$, $N_{\mathcal{I}}$ are the number of data points in $\mathcal{U}$, $\partial \mathcal{U}$ and $\Gamma$ respectively. $\phi$ is a distance function. We should note that number of Lagrange multipliers scale with the number of constraints $N_{\mathcal{B}}$, $N_{\mathcal{I}}$. When dealing with a substantial number of constraints (i.e.,$N_{\mathcal{B}}$ ,$N_{\mathcal{I}}$), the efficiency of the optimization can be compromised due to the size of the computational graph and memory requirements. To mitigate that, we propose to minimize the expected loss on the PDE while incorporating an expected equality constraint on the boundary and initial conditions. Furthermore, we justify our proposed approach by examining the distribution of the Lagrange multipliers for a large number of constraints.
% proposed formulation
%\subsection{Proposed Constrained Optimization Formulation for Solving PDEs}
\subsubsection{Computationally efficient formulation based on expectation of constraints}
\label{sec:proposed_formulation}
In this section, we introduce an efficient constrained optimization formulation, represented in the form of equation \eqref{eq:constrained_problem}, for solving partial differential equations (PDEs). In our original PECANN approach \cite{PECANN_2022}, we learned the solution of PDEs by constraining the optimization problem point-wise using the formulation described in the previous section. However, this strategy becomes computationally expensive for challenging problems, which benefit from increased number of collocation points. Constraining the learning problem pointwise is also not suitable for taking advantage of deep learning techniques designed to accelerate the training process. As we will demonstrate later in the present work, for large number of point-wise constraints, Lagrange multipliers assume a probabilistic distribution with a clear expected value. Based on this observation, we revise our formulation for point-wise constraints such that instead of constraining the loss on individual high fidelity data, we constrain the expected loss on a batch of our high fidelity data. Consequently, Lagrange multipliers for this particular approach will represent expected values, which has a direct impact on computational efficiency because Lagrange multipliers are not attached to any individual data point, we can do mini-batch training if our data does not fit in the memory for full batch training. Considering Eq.\eqref{eq:PDE} with its boundary condition \eqref{eq:BC} and initial condition \eqref{eq:IC}, we write the following constrained optimization problem: 
\begin{align}
    \min_{\theta} ~ \mathcal{J}(\theta) = \frac{1}{N_{\mathcal{F}}}\sum_{i=1}^{N_{\mathcal{F}}}\|\mathcal{F}(\boldsymbol{x}^{(i)},t^{(i)};\boldsymbol{\nu},\theta)\|_2^2,
    \label{eq:proposed_objective}
\end{align}
subject to 
\begin{align}
    \frac{1}{N_{\mathcal{B}}} &\sum_{i=1}^{N_{\mathcal{B}}} \phi( \mathcal{B}(\boldsymbol{x}^{(i)},t^{(i)},g^{(i)};\theta)) :=0,~ \forall (\boldsymbol{x}^{(i)},t^{(i)},g^{(i)}) \in \mathcal{\partial U},~ i = 1,\cdots, N_{\mathcal{B}},\\
    \frac{1}{N_{\mathcal{I}}} &\sum_{i=1}^{N_{\mathcal{I}}}\phi( \mathcal{I}(\boldsymbol{x}^{(i)},t^{(i)},h^{(i)};\theta)) :=0,~ \forall (\boldsymbol{x}^{(i)},t^{(i)},h^{(i)}) \in \Gamma,~ i = 1,\cdots, N_{\mathcal{I}},
\end{align}
where $N_{\mathcal{F}}$, $N_{\mathcal{B}}$, $N_{\mathcal{I}}$ are the number of data points in $\mathcal{U}$, $\partial \mathcal{U}$ and $\Gamma$ respectively. $\phi$ is a convex distance function. In all of the experiments $\phi$ is a quadratic function unless specified otherwise. Next, we are present a typical PDE-constrained inverse problem.

\subsubsection{Formulation for PDE-constrained inverse problems}\label{subsec:inverse_formulation}
In our original formulation of the PECANN framework \cite{PECANN_2022}, we minimized the loss on the governing PDE and noisy data while constraining the loss with high fidelity data \cite{PECANN_2022}. However, disparity in physical scales between the partial differential equations (PDEs) and noisy measurement data can exist, which couldcomplicate the inference problem. To address this challenge of solving inverse PDE problems using noisy and high fidelity data, we propose a PDE-constrained formulation for inverse problems by minimizing the loss on noisy data while considering the governing PDE and any available high fidelity data as constraints. 

Les us consider a generic inverse PDE problem to demonstrate the new formulation. We assume that the initial condition is known exactly and can be treated as high fidelity data. Additionally, we assume that the underlying physical problem is governed by a typical PDE operator as shown in Eq. \eqref{eq:PDE}.  Given a set of noisy measurement data $\{(\boldsymbol{x}^{(i)},t^{(i)}), \hat{u}^{(i)} \}_{i=1}^{N_{\mathcal{M}}}$, we can minimize the following objective
\begin{align}
    \min_{\theta,\boldsymbol{\nu}} ~ \frac{1}{N_{\mathcal{M}}}\sum_{i=1}^{N_{\mathcal{M}}}\|u(\boldsymbol{x}^{(i)},t^{(i)};\theta) - \hat{u}(\boldsymbol{x}^{(i)},t^{(i)}) \|_2^2,
    \label{eq:inverse_objective}
\end{align}
subject to 
\begin{align}
&\frac{1}{N_{\mathcal{F}}}\sum_{i=1}^{N_{\mathcal{F}}}\phi (\mathcal{F}(\boldsymbol{x}^{(i)},t^{(i)};\boldsymbol{\nu},\theta)) :=0,~ \forall (\boldsymbol{x}^{(i)},t^{(i)}) \in \mathcal{U},~ i = 1,\cdots, N_{\mathcal{F}},\\
&\frac{1}{N_{\mathcal{I}}},  \sum_{i=1}^{N_{\mathcal{I}}}\phi( \mathcal{I}(\boldsymbol{x}^{(i)},t^{(i)},h^{(i)};\theta)) :=0,~ \forall (\boldsymbol{x}^{(i)},t^{(i)},h^{(i)}) \in \Gamma,~ i = 1,\cdots, N_{\mathcal{I}},
\end{align}
where $N_{\mathcal{F}}$,  $N_{\mathcal{I}}$ are the number of data points in $\mathcal{U}$ and $\Gamma$ respectively. $\phi$ is a convex distance function. In all of the experiments $\phi$ is a quadratic function unless specified otherwise. It is worth noting that if additional high-fidelity data is available, we can incorporate it as an additional constraint into the above formulation along with other constraints.

%forward solution 
\section{Applications}\label{sec:forward_problems}
In this section, we apply our proposed augmented Lagrangian method with an adaptive update strategy to learn the solution of forward and inverse problems. We provide additional examples in the Appendix as well. Given an $n$-dimensional vector of predictions $\boldsymbol{\hat{u}} \in \mathbf{R}^n$ and an $n$-dimensional vector of exact values $\boldsymbol{u} \in \mathbf{R}^n$, we define the relative Euclidean or $\mathit{l^2}$-norm, infinity norm $\mathit{l^\infty}$-norm, and the root-mean square (RMS), respectively, to assess the accuracy of our predictions 
\begin{align}
     \mathcal{E}_r(\hat{u},u) = \frac{\|\hat{\boldsymbol{u}} - \boldsymbol{u}\|_2}{\|\boldsymbol{u}\|_2}, 
    \label{eq:relative_L2_Error} \quad
    \mathcal{E}_{\infty}(\hat{u},u) = \| \boldsymbol{\hat{u}} - \boldsymbol{u}\|_{\infty}, \quad
    \text{RMS} = \frac{1}{n}\sqrt{\sum_{i=1}^{n}(\hat{\boldsymbol{u}}^{(i)} - \boldsymbol{u}^{(i)})^2}
\end{align}
where $\|\cdot \|_2$ denotes the Euclidean norm, and $\| \cdot \|_{\infty}$ denotes the maximum norm. All the codes used in the following exampls are available as open source at \url{https://github.com/HiPerSimLab/PECANN}.

%%%% Canonical problem
\subsection{Forward problem: unsteady heat transfer in a composite medium}
In this section, we study a heat transfer problem in a composite material where temperature and heat fluxes are matched across the interface \cite{baker1985heat}. This canonical problem has initial condition, boundary condition and flux constraint which makes it ideal for highlighting the implementation intricacies of our method as well as demonstrating the significant improvement we achieve using our proposed formulation. Consider a time-dependent heat equation in a composite medium,
\begin{align}
   \frac{\partial u(x,t)}{\partial t} &= \frac{\partial }{\partial x}[\kappa(x,t) \frac{\partial u(x,t)}{\partial x}] +  s(x,t), \quad (x,t) \in \Omega \times [0,\tau]
    \label{eq:heat_pde}
\end{align}
along with Dirichlet boundary condition 
\begin{align}
    u(x,t) = g(x,t), \quad (x,t) \in \partial \Omega \times (0,\tau],
\end{align}
and initial condition 
\begin{align}
    u(x,0) = h(x),  \quad x \in \Omega,
\end{align}
where $u$ is the temperature, $s$ is a heat source function, $\kappa$ is thermal conductivity, $g$ and $h$ are source functions respectively. The composite medium consists of two non-overlapping sub-domains where $\Omega = \Omega_1 \cup \Omega_2$. We consider the thermal conductivity of the medium to vary as follows
\begin{equation}
\kappa(x,t) = 
\begin{cases} 
      1, & (x,t) \in \Omega_1 \times [0,2], \\
      3 \pi, & (x,t) \in \Omega_2 \times [0,2], 
   \end{cases}
\end{equation}
where $\Omega_1 = \{x | -1 \le x < 0 \}$ and $\Omega_2 = \{x | 0 < x \le 1 \}$. To accurately evaluate our model, we consider an exact solution of the form
\begin{equation}
u(x,t) = 
\begin{cases} 
      \sin(3 \pi  x)  t, & x \in \Omega_1 \times [0,2], \\
      t  x, & x \in \Omega_2 \times [0,2].
   \end{cases}
   \label{eq:exact_heat_solution}
\end{equation}
The corresponding source functions $s(x,t)$, $g(x,t)$ and $h(x,t)$ can be calculated exactly using \eqref{eq:exact_heat_solution}. 

First, let us introduce an auxiliary flux parameter $\sigma(x,t) = \kappa(x,t) \frac{\partial u}{\partial x}$ to obtain a system of first-order partial differential equation that reads
\begin{subequations}
\begin{align}
\mathcal{F}(x,t) &:= \frac{\partial u(x,t)}{\partial t} -  \frac{\partial \sigma(x,t) }{\partial x} +  q(x,t), \in \Omega \times [0,\tau],\\
\mathcal{Q}(x,t) &:=  \sigma(x,t) - \kappa(x,t) \frac{\partial u(x,t)}{\partial x}, \in \Omega \times [0,\tau],\\
\mathcal{B}(x,t) &:=  u(x,t) - g(x,t), \in \partial \Omega \times (0,\tau],\\
\mathcal{I}(x,t) &:=  u(x,0) - h(x), \in \Omega, t = 0,
\end{align}
\end{subequations}
where $\mathcal{F}$ is the residual form of our differential equation (used as our objective function), $\mathcal{Q}$ is the residual form of our flux operator (equality constraint), $\mathcal{B}$ is the residual form our boundary condition operator (equality constraint), and $\mathcal{I}$ is the residual form of our initial condition operator (equality constraint). 

We use a fully connected neural network architecture parameterized by $\theta$ with two inputs for $(x,t)$ and two outputs for $u$ and $\sigma$ consisting of three hidden layers with 30 neurons per layer and tangent hyperbolic activation functions. We use L-BFGS optimizer \citep{nocedal1980updating} available in the Pytorch package \cite{paszke2019pytorch} with \emph{strong Wolfe} line search function with its maximum iteration number set to five. For the purpose of this problem, we generate $N_{\mathcal{F}} = N_{\mathcal{Q}} =  10^4$ collocation points, $N_{\mathcal{B}} = 2 \times 5000$  and $N_{\mathcal{I}} = 5000$ for approximating the boundary and initial conditions only once before training. 
\begin{figure}
    \centering
    \subfloat[]{\includegraphics[scale=0.50]{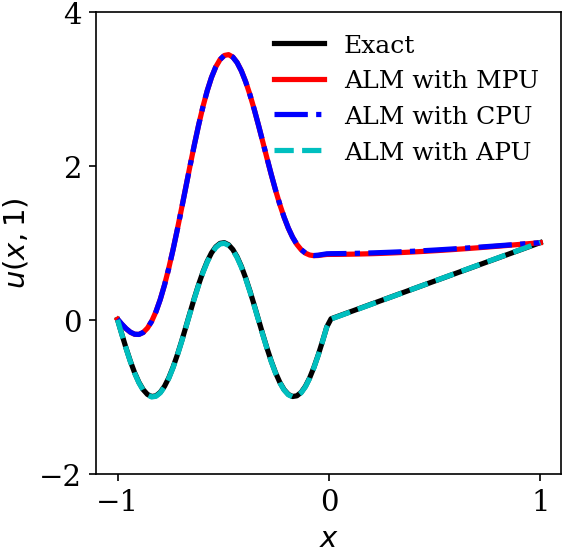}} \hspace{5em}
    \subfloat[]{\includegraphics[scale=0.50]{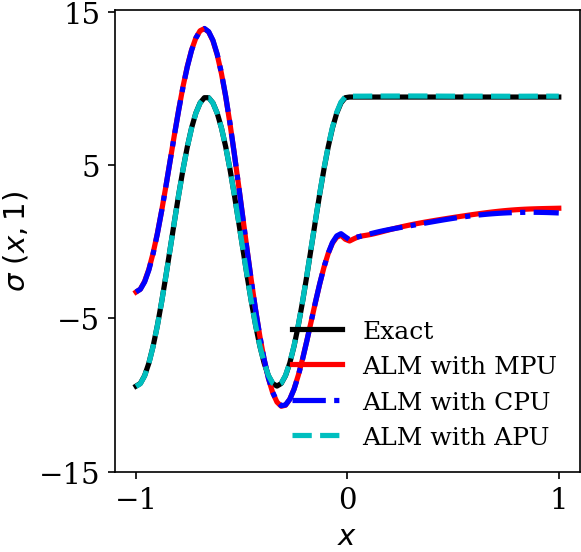}}
    \caption[Heat transfer in composite medium]{Effects of augmented Lagrangian penalty update strategies on the heat transfer in a composite medium problem using the point-wise constraint formulation (i.e. section \ref{sec:constrained_optimization_formulation_pointwise_constraint}). (a) predicted temperature at t = 1, (b) predicted heat flux at t = 1}
    \label{fig:existing_constrained_optimization_with_alm}
\end{figure}
\begin{figure}[!h]
\centering
    \subfloat[]{\includegraphics[scale=0.50]{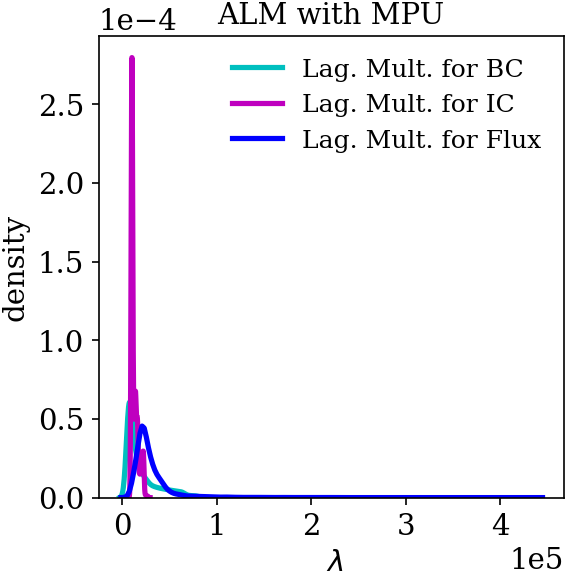}}\hspace{4em}
    \subfloat[]{\includegraphics[scale=0.50]{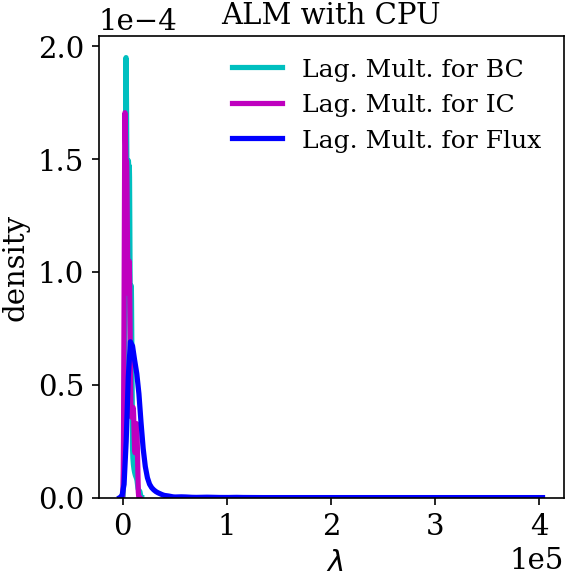}}\hspace{4em}
     \subfloat[]{\includegraphics[scale=0.50]{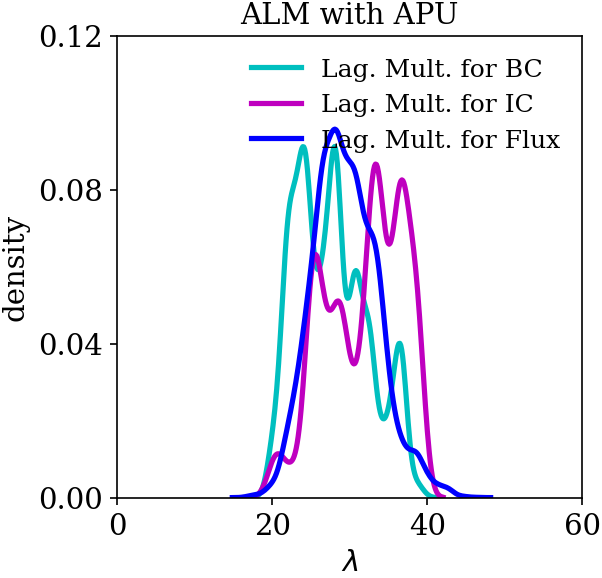}}
    \caption{Distribution of Lagrange multipliers arising from using the formulation with point-wise constraints (i.e. section \ref{sec:constrained_optimization_formulation_pointwise_constraint}). (a) ALM with monotonically increasing penalty update (Algorithm 1). (b)  ALM with conditional penalty update (Algorithm 2). (c) ALM with adaptive penalty update (Algorithm 3, proposed).
    }
\label{fig:distribution_of_lagrange_multipliers}
\end{figure}
First, we solve this problem using the formulation discussed in section \ref{sec:constrained_optimization_formulation_pointwise_constraint} using point-wise constraints. We present the results at $t=1$ obtained using the point-wise constraint formulation in Fig.\ref{fig:existing_constrained_optimization_with_alm} and compare the impact of three different strategies to update the penalty parameters in ALM. Fig.\ref{fig:existing_constrained_optimization_with_alm}(a) and (b) presents the temperature and heat flux distribution in the composite medium, respectively. We observe that adaptive penalty update strategy produces results that are in excellent agreement with the exact solution. However, results using monotonic and conditional penalty updates produce a solution that is markedly different than the exact solution.

With Fig.\ref{fig:existing_constrained_optimization_with_alm} we show the importance of adaptively updating the penalty parameter when dealing with diverse constraint types (e.g. boundary condition, initial confition, and flux conservation) that exhibit different physical characteristics. However, we apply the constraints point-wise. As we mentioned previously, the computational cost  of this particular implementation scales linearly with the number of points assigned to each constraint, especially when constraining the flux because it is applied to every collocation point in the domain. Furthermore, with the point-wise constraint formulation, we cannot train our model in mini-batch training fashion since each Lagrange multiplier is attached to every constraint data point. 

In Fig.~\ref{fig:distribution_of_lagrange_multipliers}, we present the distribution of Lagrange multipliers resulting from the point-wise constraint formulation presented in section \ref{sec:constrained_optimization_formulation_pointwise_constraint}. Our goal is to find out if it is necessary to constrain the solution point-wise. We observe that the distribution of Lagrange multipliers assumes probabilistic distributions regardless of the penalty update strategy. This observation indicates that when there are a large number of point-wise constraints, we can constrain an expected value of the loss on each type of constraints using a single Lagrange multiplier as seen in Figs.~\ref{fig:distribution_of_lagrange_multipliers}(a)-(c). We should note that expected loss is also commonly used as an objective function in machine learning applications. Hence, we employ our proposed formulation to train our models using ALM with different penalty updating approaches to showcase the efficiency of our adaptive ALM and our proposed formulation. 

\begin{figure}
\centering
    \subfloat[]{\includegraphics[scale=0.50]{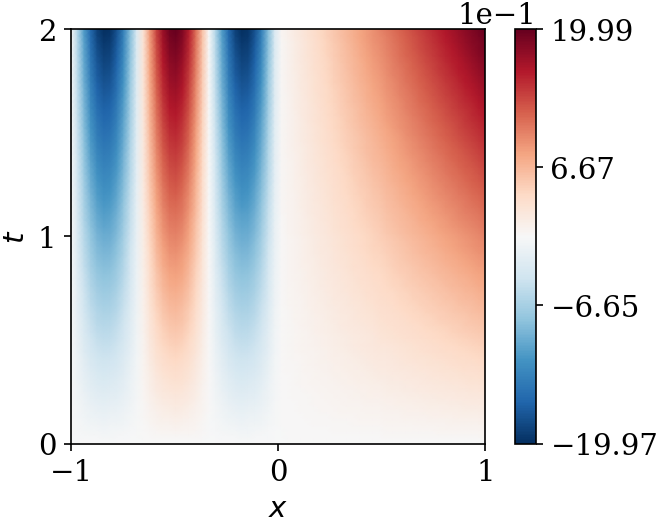}}\hspace{4em}
    \subfloat[]{\includegraphics[scale=0.50]{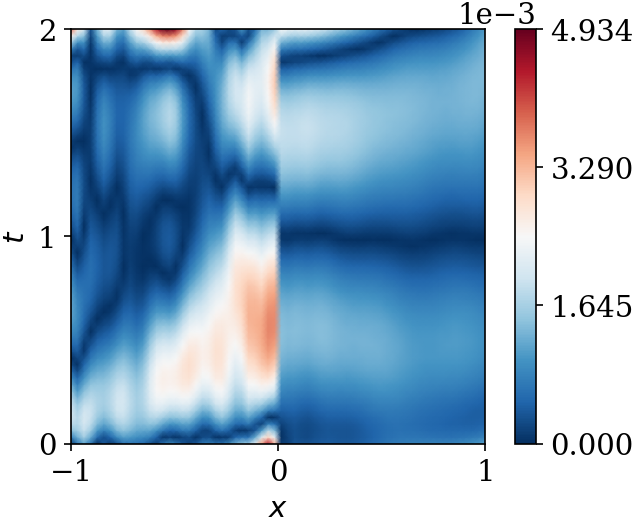}}\\
    \subfloat[]{\includegraphics[scale=0.50]{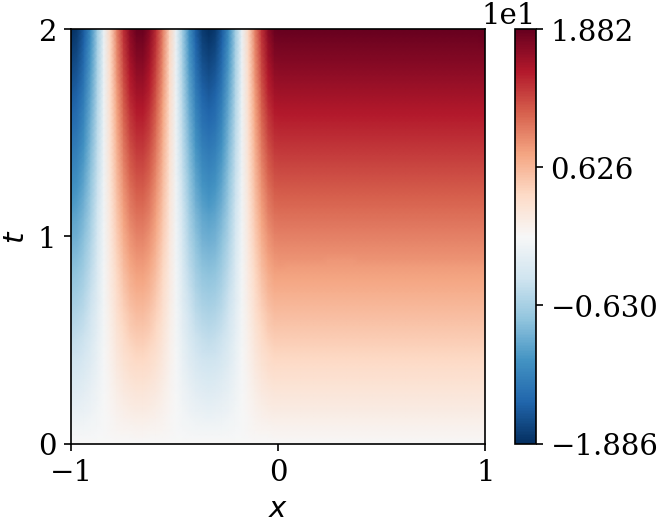}}\hspace{4em}
    \subfloat[]{\includegraphics[scale=0.50]{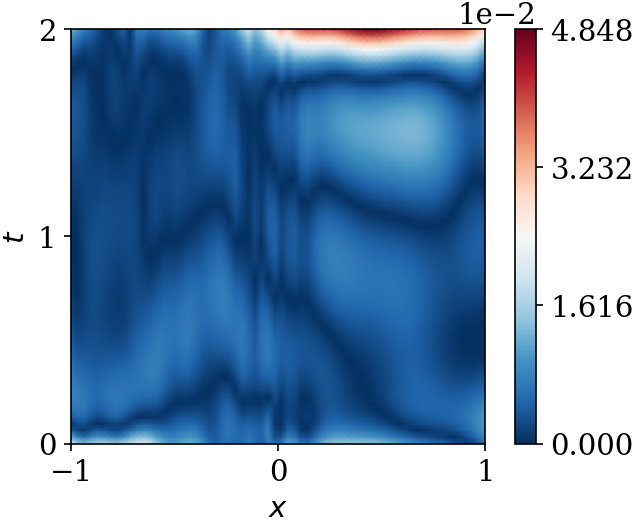}}
    
    \caption{Heat transfer in a composite medium using the adaptive penalty update strategy and the formulation based expectation of constraints (i.e. section \ref{sec:proposed_formulation}). (a) predicted temperature (b) absolute point-wise error in predicted temperature , (c) predicted heat flux, (d) absolute point-wise error in heat flux prediction.}
    \label{fig:expected_global_constraint}
\end{figure}

\begin{figure}
\centering
    \subfloat[]{\includegraphics[scale=0.50]{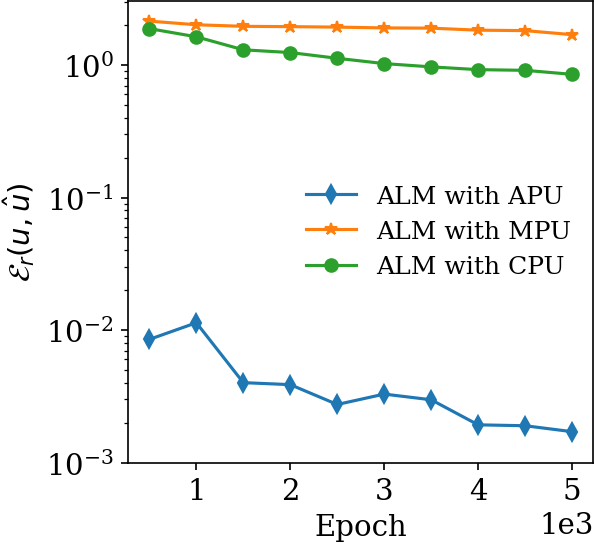}}\hspace{4em}
    \subfloat[]{\includegraphics[scale=0.50]{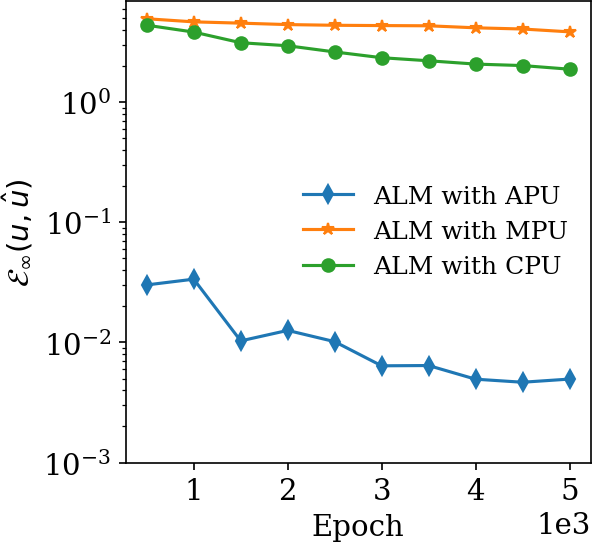}}\\
    \subfloat[]{\includegraphics[scale=0.50]{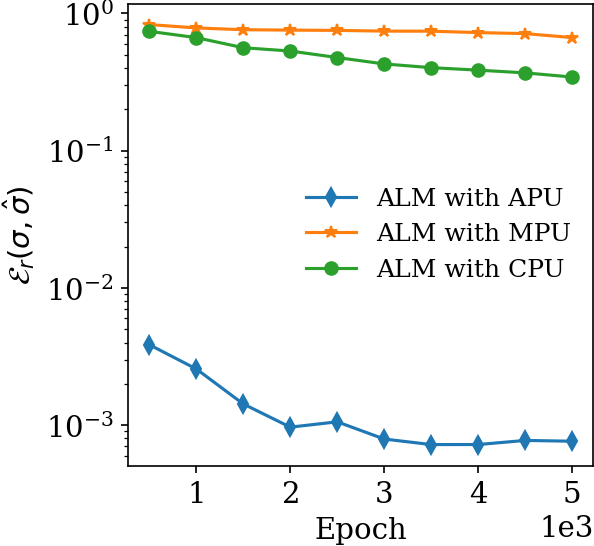}}\hspace{4em}
    \subfloat[]{\includegraphics[scale=0.50]{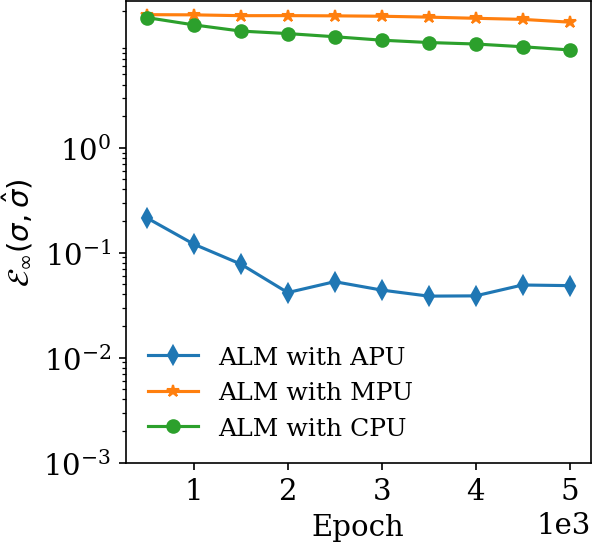}}
    \caption{Error levels during training using ALM with different penalty update strategy. Top row: Temperature (a) relative error (b) absolute error. Bottom row: Heat flux (c) relative error, (d) absolute error.}
    \label{fig:error_levels}
\end{figure}

Figure \ref{fig:expected_global_constraint} presents the results of our PECANN model with the adaptive ALM and with the expected loss formulation given in section \ref{sec:proposed_formulation} applied to the heat transfer in a composite medium problem. Specifically, Fig.\ref{fig:expected_global_constraint}(a)-(b) presents the space and time solution of the temperature and the absolute value of the point-wise error in temperature produced by our model, respectively. Space and time solution of the heat flux and its absolute value error are shown in Fig.\ref{fig:expected_global_constraint}(c)-(d), respectively. Collectively, Fig. \ref{fig:expected_global_constraint} demonstrates our predictions closely match the exact solutions. Our PECANN model using the monotonic (MPU) and conditional (CPU) penalty update strategies do not converge, and therefore we do not report temperature and heat flux predictions using those two strategies. However, we provide the error norms resulting from all three penalty update strategies for comparison purposes in Fig.~\ref{fig:error_levels} from which we observe that ALM with adaptive penalty update strategy outperforms the other two strategies. From Fig.~\ref{fig:error_levels}, we also observe that adaptive penalty update (APU) strategy can provide acceptable error levels with even smaller number of epochs. 

\begin{figure}
\centering
    \subfloat[]{\includegraphics[scale=0.50]{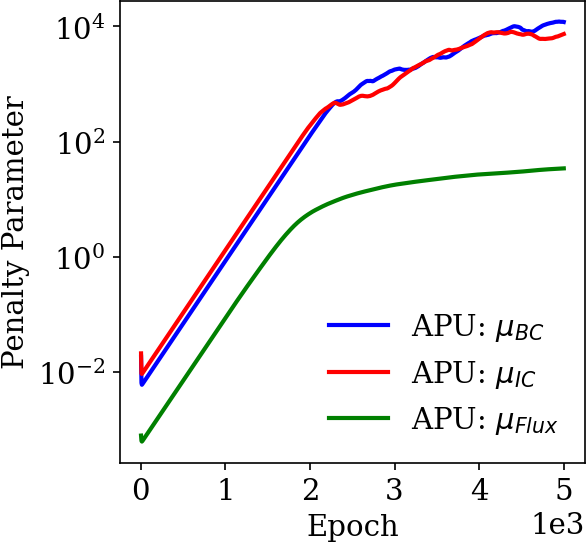}} \hspace{6em}
     \subfloat[]{\includegraphics[scale=0.50]{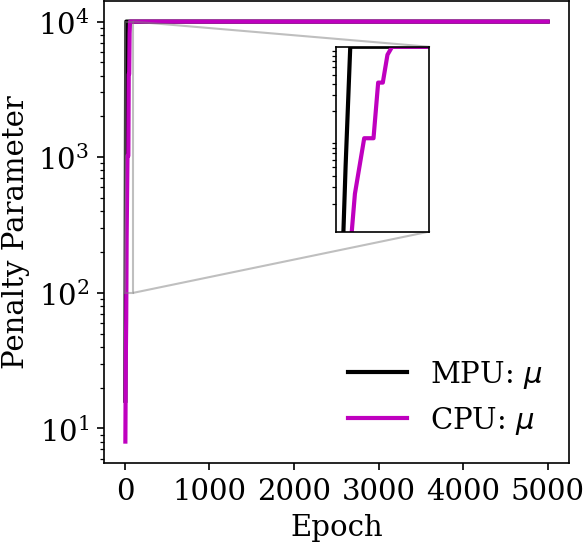}} \\
    \subfloat[]{\includegraphics[scale=0.55]{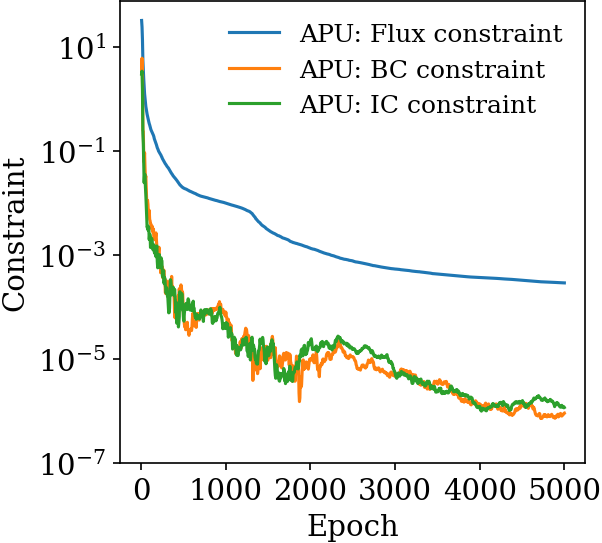}}   
    \caption[Evolution of penalty parameters with different penalty update strategy]{Evolution of penalty parameters with different penalty update strategy: (a) evolution of the penalty parameter $\mu_{\mathcal{B}}$ for enforcing boundary conditions, $\mu_{\mathcal{I}}$ for enforcing initial conditions, and $\mu_{\mathcal{Q}}$ for enforcing flux constraints using ALM with adaptive penalty update strategy . (b) evolution of penalty parameters for enforcing boundary condition, initial condition and flux equation using ALM with MPU and CPU. (c) evolution of constraints during training using ALM with APU}
\label{fig:penalty_parameters}
\end{figure}

To further gain insight into how ALM with adaptive penalty updates (APU) works, we analyze the evolution of the penalty parameters during training. Fig.\ref{fig:penalty_parameters}(a) shows the evolution of penalty parameters with each epoch for enforcing boundary conditions $\mu_{BC}$, initial conditions $\mu_{IC}$, and flux constraints $\mu_{Flux}$ using the adaptive penalty update (APU). The penalty parameters for each constraint type grow at different rates over the course of training, justifying the need for adaptive and independent penalty parameters for each constraint type. Fig.\ref{fig:penalty_parameters}(b) illustrates the evolution of the penalty parameter for enforcing boundary conditions, initial conditions, and flux constraints using conditional penalty update (CPU) and monotonically increasing penalty update (MPU) strategies. In both cases, the penalty parameter quickly reaches the maximum penalty value of $\mu_{\max} = 10^{4}$, causing the problem to become ill-conditioned and leading to divergence during optimization. This observation again highlights the importance of using an adaptive penalty update strategy to avoid ill-conditioning and ensure convergence. Additionally, in Fig.\ref{fig:penalty_parameters}(c), we present the evolution of the loss on our equality constraints during training, when using the adaptive penalty update strategy. We observe that all our constraints decay at different rates, which provides further justification for the varying rates at which the penalty parameters increase.

Our investigation and findings from the heat transfer in a composite medium problem provide a strong support for our proposed augmented Lagrangian method with an adaptive penalty update strategy (i.e. Algorithm \ref{alg:adaptive_training_algorithm}) and for the computationally efficient formulation of the objective function using expectation of constraints (i.e. formulation given in section \ref{sec:proposed_formulation}). In the rest of the examples, we will adopt this combination to further showcase its efficacy and versatility.  

%%%
\subsection{Forward problem: wave equation}
Baseline PINN model \cite{raissi2017physicsII} with a fully connected neural network architecture is known to struggle for certain type of PDEs \cite{mcclenny2020self, wang_parisi_tangent_kernel, basir2022critical}. \citet{wang_parisi_tangent_kernel} and \citet{mcclenny2020self} single out the one-dimensional (1D) wave equation as a challenge case for future PINN models because the baseline PINN model faces severe challenges when applied to this PDE problem. Here, we apply our PECANN model with the proposed adaptive ALM to learn the solution of the same 1D wave equation as studied in \cite{wang_parisi_tangent_kernel, mcclenny2020self}  and compare the accuracy of our predictions and the size of our neural network model against the results and networks presented in those two studies. While comparing the accuracy levels is significant, we also believe it is imperative to take into perspective the size of the neural network architecture, and the number of the collocations points deployed to obtain the level of accuracy in predictions.

Consider the following 1D wave equation
\begin{align}
    &\frac{\partial^2 u}{\partial t^2} - 4 \frac{\partial^2 u}{\partial x^2}  = 0, \quad (x,t) \in \mathcal{U},
    \label{eq:wave_pde}
\end{align}
with the boundary condition
\begin{align}
    &u(x,t) = 0, \quad (x,t) \in \partial \mathcal{U},
    \label{eq:wave_bc}
\end{align}
and the initial condition 
\begin{align}
        \frac{\partial u}{\partial t}(x,t) = 0, \quad (x,t) \in \partial \mathcal{U}.    \label{eq:wave_ic}
\end{align}
The exact solution to the problem is
\begin{align}
    u(x,t) = \sin(\pi x) + \frac{1}{2}\sin(4 \pi x ) , \quad (x,t) \in \mathcal{U},
\end{align}
where $\mathcal{U} = \{(x,t) ~|~x \in \Omega, t = [0,1] \}$,
$\partial \mathcal{U} = \{(x,t) ~|~ x \in \partial \Omega, t = [0,1]\}$ and $\Gamma = \{(x,t) ~|~x \in \Omega, t = 0\}$. 
$\partial \Omega$ represents the boundary of the spatial domain $\Omega = \{ x | x = (0,1) \}$.

\begin{table}[b]
\centering
\caption{Comparison of relative error $\mathcal{E}_r$ levels and network size used in the solution of 1D wave equation.}
\label{tb:wave}
\vspace{2pt}
\resizebox{0.95\textwidth}{!}{%
\begin{tabular}{@{}rlcccccr@{}}
\toprule
\multicolumn{1}{c}{Models} &
  \multicolumn{1}{c}{Mean $\mathcal{E}_r(u,\hat{u}) \pm$ stdev; (trials)} &
  \multicolumn{1}{c}{No. trials } &
  \multicolumn{1}{c}{$H_L \times N_H$ } &
    \multicolumn{1}{c}{$N_{\mathcal{U}}$} &
  \multicolumn{1}{c}{$N_{\partial \mathcal{U}}$}&
  \multicolumn{1}{c}{$N_{\Gamma}$}&
\\ \midrule
Baseline PINN \citep{wang_parisi_tangent_kernel} & $4.518 \times 10^{-1}$ & $1$& $5 \times 500$& $24 \times 10^6$ &$24 \times 10^6$ &$24 \times 10^6$& \\
PINN using NTK \citep{wang_parisi_tangent_kernel} & $1.728 \times 10^{-3}$ &$1$& $5 \times 500$& $24\times 10^6$ &$24 \times 10^6$ &$24 \times 10^6$&\\
Baseline PINN\citep{mcclenny2020self} & $3.792 \times 10^{-1} \pm 0.162 \times 10^{-1}$ & $10$&$5 \times 500$& $24 \times 10^6$ &$8 \times 10^6$ &$8 \times 10^6$&\\
SA-PINN \citep{mcclenny2020self} & $8.105 \times 10^{-1} \pm 1.591 \times 10^{-1}$ & $10$&$5 \times 500$& $24 \times 10^6$ &$8 \times 10^6$ &$8 \times 10^6$&\\
SA-PINN with SGD \citep{mcclenny2020self} & $2.950 \times 10^{-2} \pm 7.0 \times 10^{-3}$ & $10$&$5 \times 500$& $24 \times 10^6$ &$8 \times 10^6$ &$8 \times 10^6$&\\
\textbf{Current Method }& $\boldsymbol{3.990 \times 10^{-3} \pm 9.179 \times 10^{-4} }$& $10$&$\boldsymbol{1 \times 50}$& $\boldsymbol{300}$ &$\boldsymbol{300}$ &$\boldsymbol{300}$&
\end{tabular}}
\end{table}
\citet{wang_parisi_tangent_kernel, mcclenny2020self} used a fully connected feed-forward neural network with five hidden layer $(H_L = 5)$ and 500 neurons per hidden layer $(N_H = 500)$ with $N_{\mathcal{U}} = 300$ collocation points, $N_{\mathcal{\partial U}} = 300 $ boundary points and $N_{\Gamma} = 300$ initial points that are generated at every epoch to solve the same 1D wave equation. In the present work, we use a single hidden layer $(H_L = 1)$ with 50 neurons $(N_H = 50) $ and a tangent hyperbolic activation function. We sample $N_{\mathcal{U}} = 300$ collocation points, $N_{\mathcal{\partial U}} = 300 $ boundary points and $N_{\Gamma} = 300$ only once before training. Our objective function is the residual on the governing PDE given in \eqref{eq:wave_pde} and our constraints are the boundary condition \eqref{eq:wave_bc} and initial conditions \eqref{eq:wave_ic}. We assign three Lagrange multipliers and three penalty parameters for this problem. We train our model for $10,000$ epochs even though we obtain an acceptable level of accuracy with $2,000$ epochs. We use L-BFGS optimizer with its default parameters in the Pytorch package. Table \ref{tb:wave} presents a summary of the prediction errors, architecture size and collocation points for this problem in comparison to other works. 

\begin{figure}
    \centering
    \subfloat[]{\includegraphics[scale=0.4]{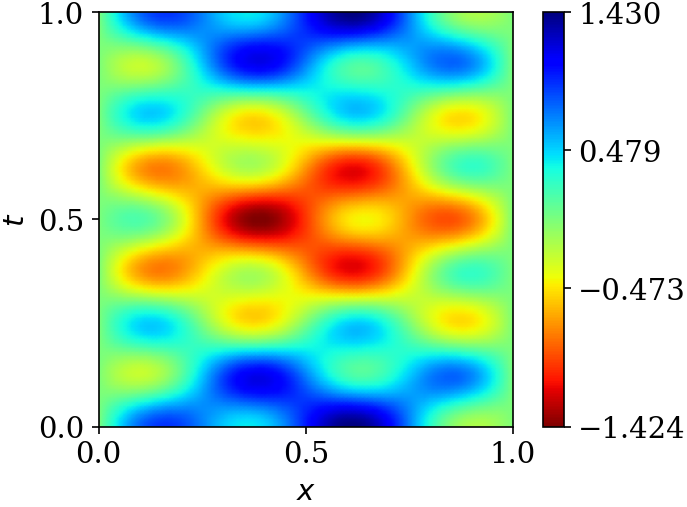}}\qquad
    \subfloat[]{\includegraphics[scale=0.4]{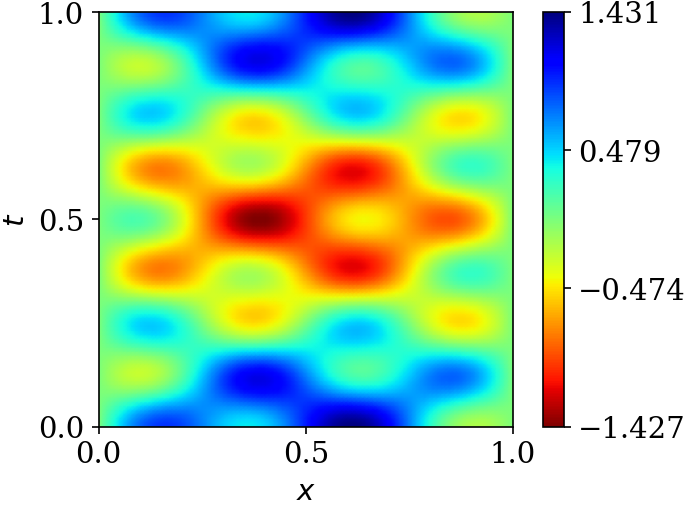}}\qquad
    \subfloat[]{\includegraphics[scale=0.4]{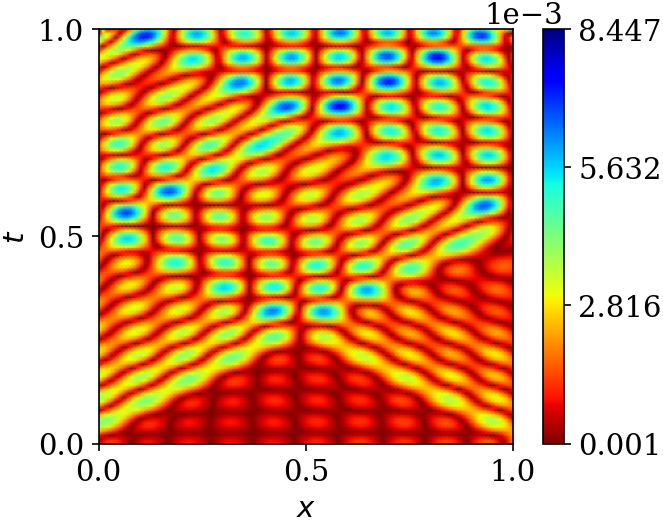}}
    \caption{Solution of 1D wave equation obtained from a single hidden layer neural network with 50 neurons trained using the proposed augmented Lagrangian method: (a) predicted solution, (b) exact solution (c) absolute pointwise error }
    \label{fig:wave_eq}
\end{figure}

We present the prediction of our model in Fig.~\ref{fig:wave_eq}.
We can observe that the our model has successfully learned the underlying solution to a good level of accuracy. In Table \ref{tb:wave} we compare our predictions against other works that tackled the same problem. We report the mean and standard deviation of the relative error calculated over 10 different trials. The results show that our method produces results with error levels that are one order of magnitude level than the results obtained with the self-adaptive PINNs \cite{mcclenny2020self}, while using a much smaller neural network model.

% incompressible navier-stokes
\subsection{Forward problem: lid-driven cavity flow simulation up to a Reynolds number of 1000}\label{chapt:incompressible_fluid_flows}
In this section, we apply our adaptive augmented Lagrangian method to solve the steady-state flow in a two-dimensional lid-driven cavity benchmark problem \cite{ghia_1982} at Reynolds numbers of 100, 400, and 1000. Unlike other works where a stream function vorticity formulation was adopted, we use the primitive-variables formulation of the Navier-Stokes equations, which potentially makes our approach extensible to three-dimensional fluid flow problems. We should also mention that lid-driven cavity at Reynolds of 1000 is considered much more challenging than the same problem at lower Reynolds numbers, such as at $Re=100$ or $Re=400$. 

\begin{figure}[!h]
 \centering
  \includegraphics[scale=0.50]{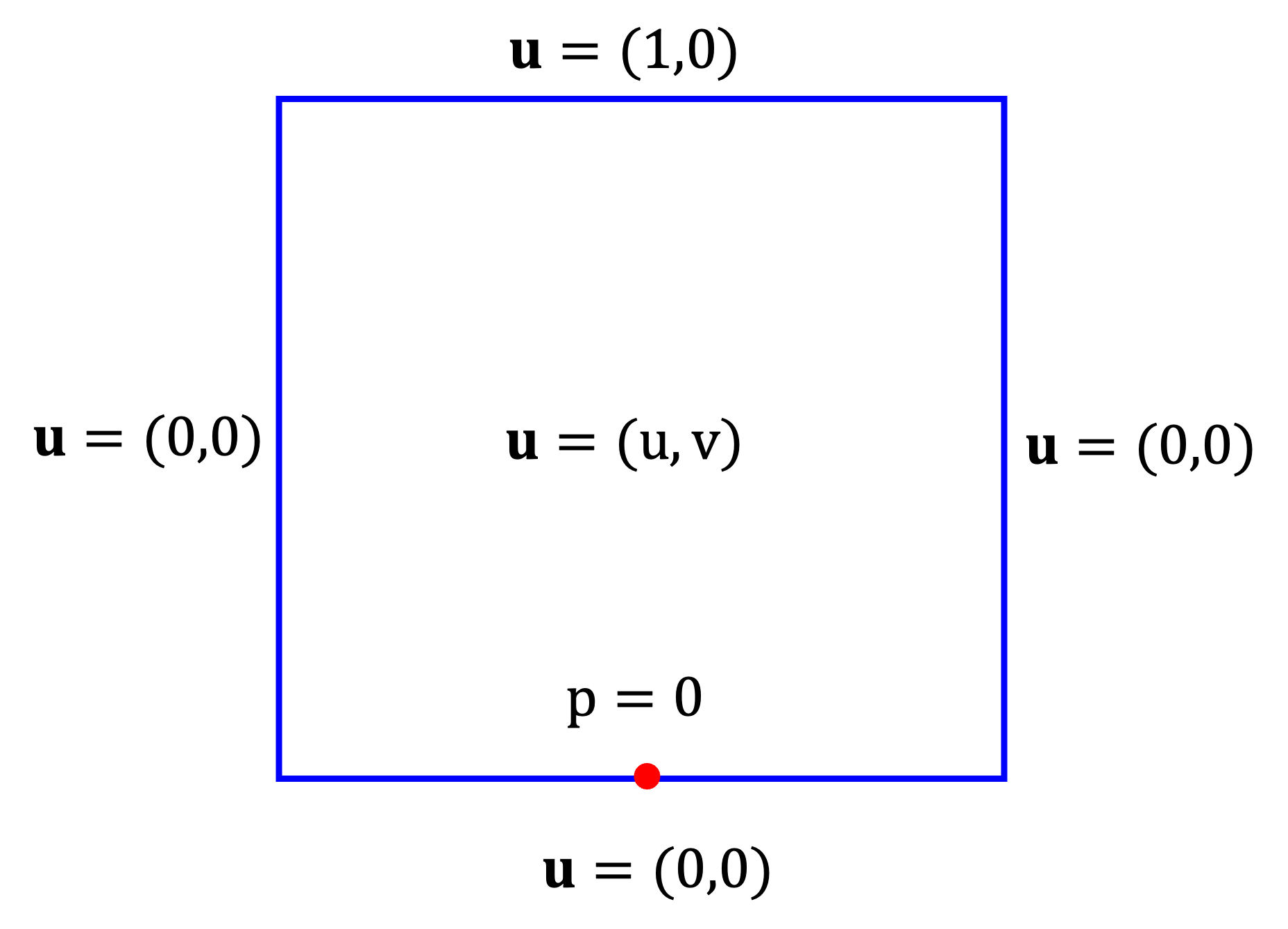}
\caption{Boundary conditions for the lid-driven cavity problem}
\label{fig:proposed_navier_stokes_domain}
\end{figure}

Steady-state Navier-Stokes equations in two dimensions and the continuity equation are written as follows: 
\begin{subequations}
\begin{align}
    u \frac{\partial u}{\partial x} + v \frac{\partial u}{\partial y} + \frac{\partial p}{\partial x} - \frac{1}{Re}(\frac{\partial^2 u}{\partial x^2} + \frac{\partial^2 u}{\partial y^2}) &=0, \quad (x,y) \in \Omega\\
    u \frac{\partial v}{\partial x} + v \frac{\partial v}{\partial y} +  \frac{\partial p}{\partial y} - \frac{1}{Re}(\frac{\partial^2 v}{\partial x^2} + \frac{\partial^2 v}{\partial y^2}) &= 0,\quad (x,y) \in \Omega,\\
    \frac{\partial u}{\partial x} + \frac{\partial v}{\partial y} &= 0 , \quad (x,y) \in \Omega 
\end{align}
\label{eq:Lid_driven_cavity}
\end{subequations}
where $\boldsymbol{u} = (u,v)$ is the velocity vector, $Re$ is the Reynolds number and $p$ is the pressure. We aim to learn the solution of the momentum equations constrained by the continuity equation on $\Omega = \{(x,y)~|~ 0 \le x \le 1, 0 \le y \le 1 \}$ with its boundary $\partial \Omega$. No-slip boundary conditions are imposed on all walls of the square cavity. The top wall of the square cavity moves at a velocity of $u(x,1) = 1$ in the $x$ direction. Because we have a closed system with no inlets or outlets in which the pressure level is defined, we fix the pressure field at an arbitrary reference point as $p(0.5,0) = 0$. Fig.\ref{fig:proposed_navier_stokes_domain} illustrates our simulation setup for the lid-driven cavity flow problem. We should note that our objective function in this problem is the loss on the total momentum $\mathcal{F}$ field, and our constraints are the continuity equation (i.e. divergence-free velocity field) $\mathcal{Q}$, boundary conditions and the pressure that is fixed at a reference point.  

To formulate our constrained optimization problem,
let us define $\boldsymbol{\mathcal{F}} = (\mathcal{F}_x,\mathcal{F}_y)$, which represents the steady-state, two-dimensional momentum equations, and $\mathcal{Q}$ as the divergence-free condition as follows
\begin{subequations}
\begin{align}
    \mathcal{F}_x(x,y) &:= u \frac{\partial u}{\partial x} + v \frac{\partial u}{\partial y} + \frac{\partial p}{\partial x} - \frac{1}{Re}(\frac{\partial^2 u}{\partial x^2} + \frac{\partial^2 u}{\partial y^2}) , \quad (x,y) \in \Omega\\
    \mathcal{F}_y(x,y) &:=u \frac{\partial v}{\partial x} + v \frac{\partial v}{\partial y} +  \frac{\partial p}{\partial y} - \frac{1}{Re}(\frac{\partial^2 v}{\partial x^2} + \frac{\partial^2 v}{\partial y^2}),\quad (x,y) \in \Omega,\\
    \mathcal{Q}(x,y) &:= \frac{\partial u}{\partial x} + \frac{\partial v}{\partial y} , \quad (x,y) \in \Omega
\end{align}
\end{subequations}

We sample $N_{\mathcal{B}} = 4 \times 64$ points for approximating the boundary conditions, $N_{\mathcal{Q}} = N_{\mathcal{F}} = 10^4$ points for approximating the loss on conservation of mass and momentum only once before training, respectively. Additionally, we constrain a single point for pressure as shown in Fig.\ref{fig:proposed_navier_stokes_domain}. We use a fully connected neural network architecture consisting of four hidden layers with 30 neurons in each layer and the tangent hyperbolic activation function. We choose L-BFGS optimizer with its default parameters and \emph{strong Wolfe} line search function that are available in the PyTorch package. We train our network for 2000 epochs for each Reynolds number separately.

\begin{figure}[!h]
 \centering
    %\subfloat[]{\includegraphics[scale=0.55]{figures/NS/driven_flow_augmented_lagrangian_Re_100.png}}\quad
    \subfloat[]{\includegraphics[scale=0.50]{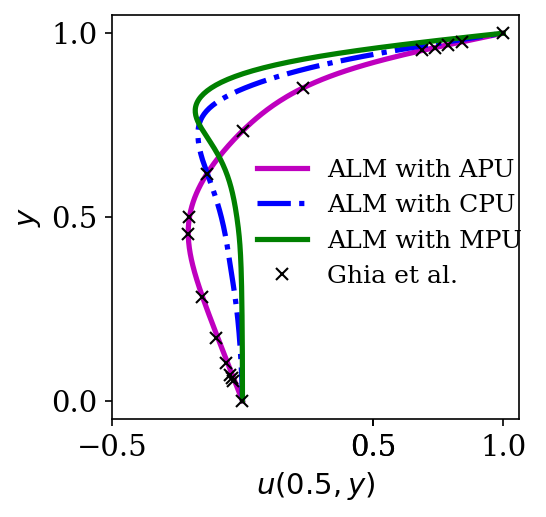}}\quad
    \subfloat[]{\includegraphics[scale=0.50]{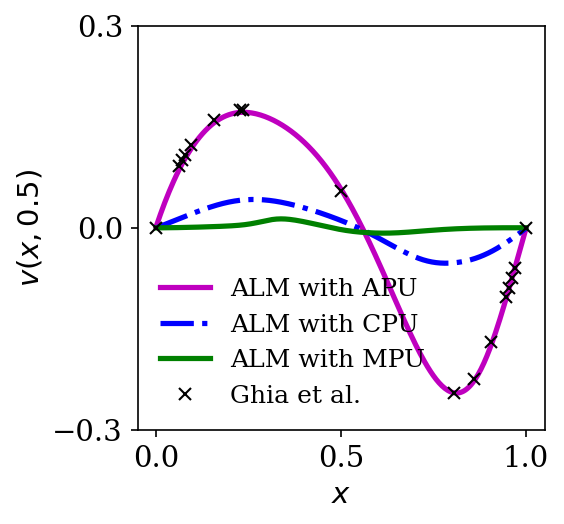}} 
    \caption{Impact of three different ALM penalty update strategies on the solution of lid-driven cavity problem at Re = 100: (a) predicted horizontal velocity along the line at $x=0.5$ compared against the benchmark data, (b) predicted vertical velocity along the line at $y=0.5$ compared against the benchmark data.}
    \label{fig:ldc_re_100}
\end{figure}
In Fig.~\ref{fig:ldc_re_100}, we investigate the effect of different penalty update strategies on the predictions. We observe that our model trained with the adaptive penalty update strategy in ALM learns the underlying solution for Re=100 well and matches the benchmark data closely. Figs.~\ref{fig:ldc_re_100}(a)-(b) also show that ALM with monotonic (MPU) or conditional (CPU) penalty update strategies are incapable of producing the expected solution. Next, we consider the case with Reynolds number 400 and train our model for 2000 epochs using the adaptive penalty update strategy only since the other two strategies did not produce acceptable results for $Re=100$. 
\begin{figure}[!h]
 \centering
    %\subfloat[]{\includegraphics[scale=0.67]{figures/NS/driven_flow_augmented_lagrangian_Re_400.png}}
    \subfloat[]{\includegraphics[scale=0.60]{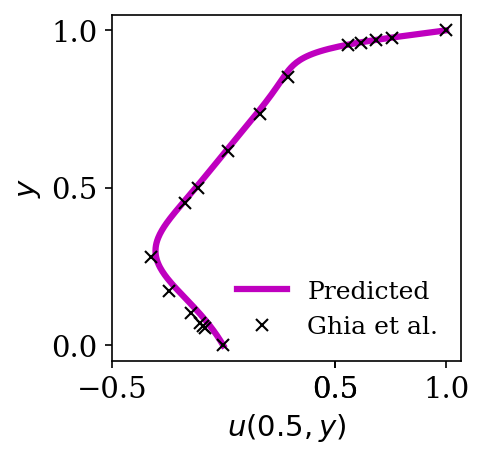}}
    \subfloat[]{\includegraphics[scale=0.60]{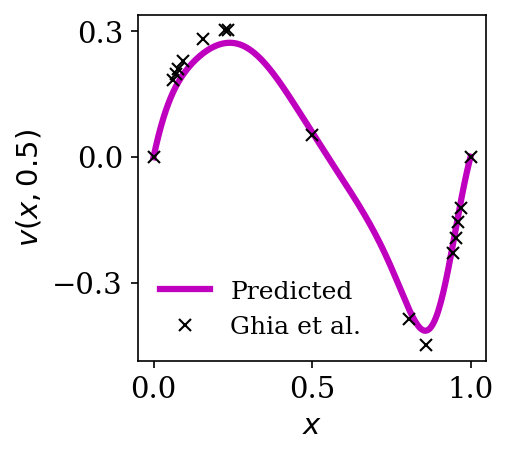}}
    \caption{Solution of lid-driven cavity problem at Re = 400 using ALM with APU: 
    %(a) predicted velocity streamlines, 
    (a) predicted horizontal velocity along the line at $x=0.5$ compared against the benchmark data, (b) predicted vertical velocity along the line at $y=0.5$ compared against the benchmark data.}
    \label{fig:ldc_re_400}
\end{figure}

\begin{figure}[!h]
 \centering
    %\subfloat[]{\includegraphics[scale=0.67]{figures/NS/driven_flow_augmented_lagrangian_Re_1000.png}}
    \subfloat[]{\includegraphics[scale=0.60]{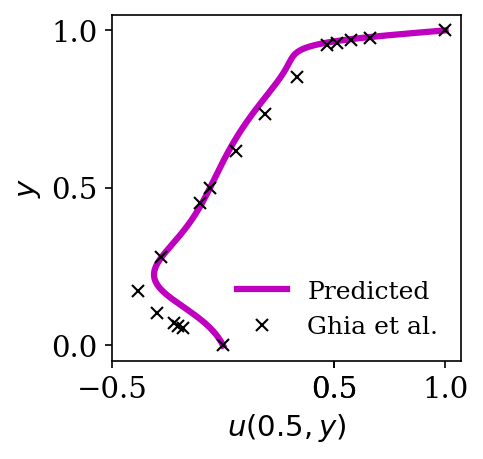}}
    \subfloat[]{\includegraphics[scale=0.60]{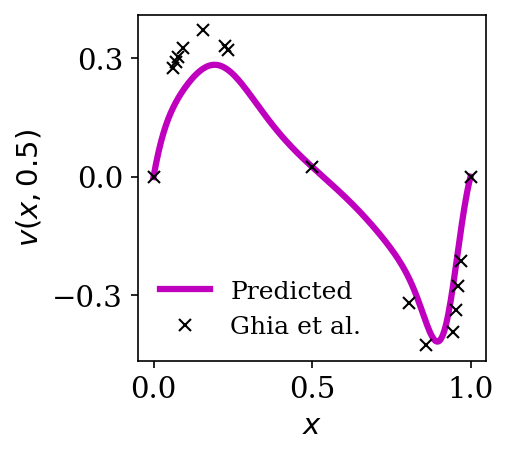}} 
    \caption{ Same caption as in Fig.~\ref{fig:ldc_re_400}, but for $Re=1000$
    %Solution of lid-driven cavity problem at Re = 1000 (a) predicted velocity streamlines, (b) predicted horizontal velocity over line compared with the benchmark result, (c) predicted vertical velocity over line compared with the benchmark result
    }
    \label{fig:ldc_re_1000}
\end{figure}

We observe from Fig.~\ref{fig:ldc_re_400} and Fig.~\ref{fig:ldc_re_1000} that our model with the adaptive update strategy learns the underlying pattern for $Re=400$ and $Re=1000$, respectively. However, the predicted velocity fields do not closely match the benchmark numerical solution, despite the our model's superior performance for the $Re=100$ case. 

It is worth noting that our objective in these two simulation cases is to highlight the fact that despite implementing and constraining appropriate boundary conditions, divergence-free velocity field, and anchoring the pressure at a single point, the task of learning becomes increasingly more challenging as the Reynolds number increases. 
%Next, we further increase the Reynolds number to 1000, which is considered the most challenging case for numerical methods when simulating the lid-driven cavity benchmark problem.
\begin{figure}[!h]
 \centering
      \subfloat[]{\includegraphics[scale=0.60]{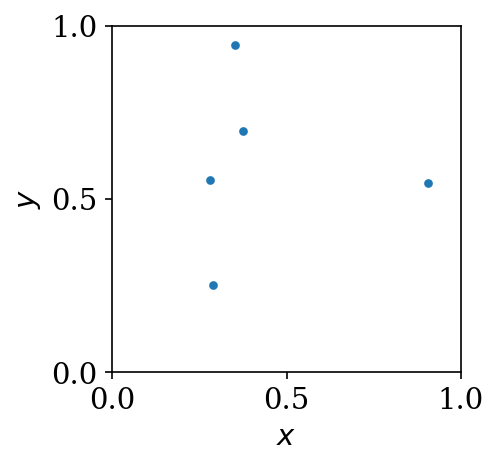}}\qquad
    \subfloat[]{\includegraphics[scale=0.60]{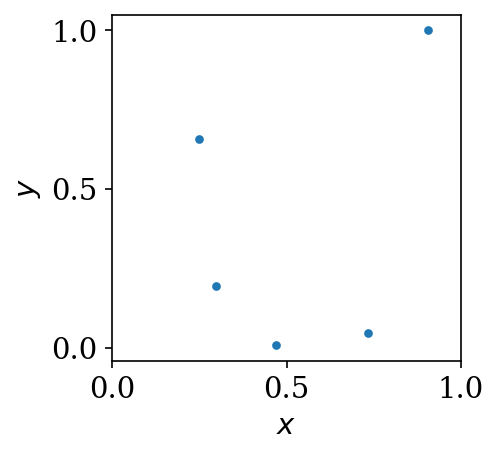}}\\
     \subfloat[]{\includegraphics[scale=0.67]{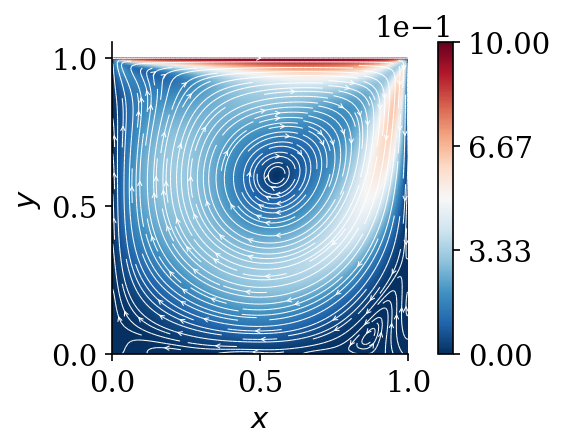}}
    \subfloat[]{\includegraphics[scale=0.60]{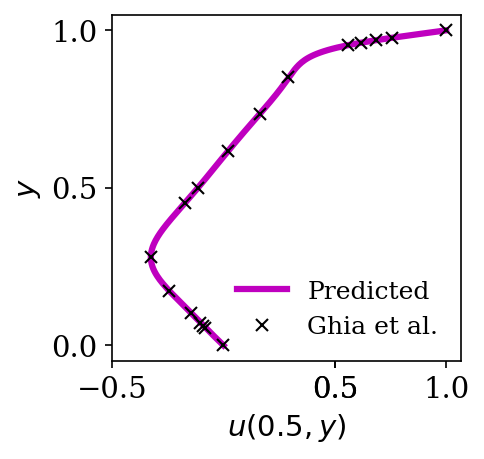}}
    \subfloat[]{\includegraphics[scale=0.60]{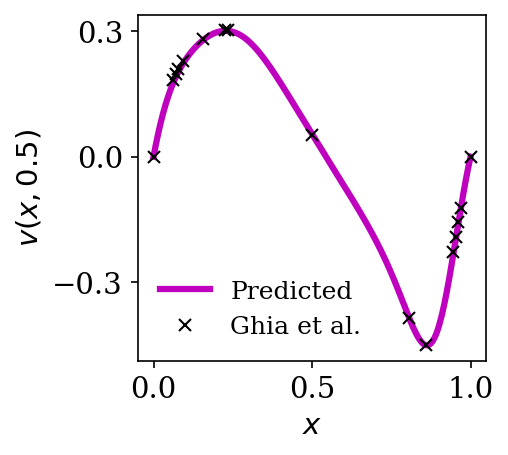}} \\
    \subfloat[]{\includegraphics[scale=0.67]{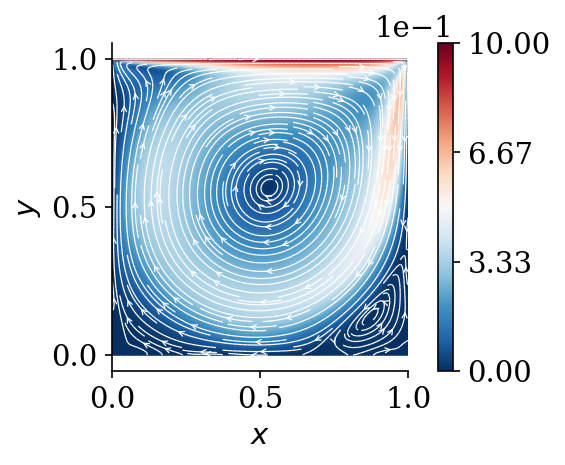}}
    \subfloat[]{\includegraphics[scale=0.60]{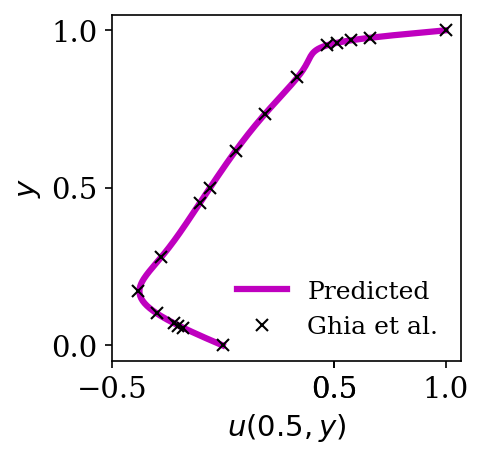}}
    \subfloat[]{\includegraphics[scale=0.60]{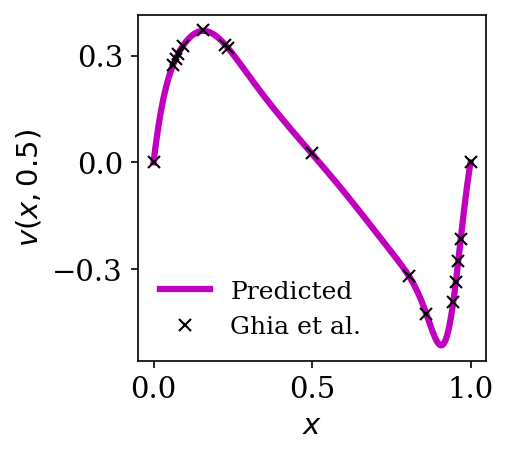}} 
    \caption{Lid-driven cavity predictions with a small sample of high fidelity velocity data for different Reynolds numbers. ALM with APU is used in the computations. Locations of randomly sampled high fidelity data:  (a) Re = 400, (b) Re = 1000. Results for Re=400: (c) predicted velocity streamlines, (d) predicted horizontal velocity along the line at $x=0.5$ compared against the benchmark result, (e) predicted vertical velocity along the line at $y=0.5$ compared against the benchmark result. Bottom row: Results for Re = 1000, same captions as in (c)-(e) 
    %(f) predicted velocity streamlines, (g) predicted horizontal velocity over line compared with the benchmark result, (h) predicted vertical velocity over line compared with the benchmark result.
    }
    \label{fig:ldc_data_driven_1000}
\end{figure}
Learning the solution of lid-driven cavity problem at a high Reynolds numbers may be achievable with a larger neural network or a higher number of collocations. However, our aim here is not to demonstrate whether such learning is possible or not. Instead, we aim to illustrate the challenges involved for high Reynolds number flow problems, even when formulated properly as a constrained optimization problem. We hypothesize that for high Reynolds number flow problems, the underlying optimization problems become ill-conditioned and require additional regularization. Therefore, in order to further regularize the optimization process, we introduce a small set of exact, ramdomly sampled velocity data within the flow domain as shown in Fig. \ref{fig:ldc_data_driven_1000}(a)-(b). We use these clean data points as equality constraints on the prediction of our model, similar to the way we handle boundary conditions.

We observe from Figs.~\ref{fig:ldc_data_driven_1000}(c)-(e) that our PECANN model with an adaptive ALM successfully predicts the fluid flow in the cavity at Reynolds number = 400 with a high degree of accuracy by leveraging a small sample of high fidelity data. Specifically, Figs.~\ref{fig:ldc_data_driven_1000}(c)-(e) show the predicted solution, x-component of the velocity over a line at $x=0.5)$, and y-component of the velocity over a line at $y=0.5)$ for $Re=400$, while Figs.~\ref{fig:ldc_data_driven_1000}(f)-(h) show the corresponding results for $Re=1000$. Compared to the results shown in Fig.~\ref{fig:ldc_re_400} and Fig.~\ref{fig:ldc_re_1000}, our new results suggest that a small set of randomly sampled labeled data is beneficial to regularize the optimization process for high Reynolds numbers. We aim to explore the feasibility of learning high Reynolds numbers in two and three dimensions without any labeled data using larger and different neural network architectures in the future.

\subsection{Inverse problem: estimating transient thermal boundary condition from noisy data}\label{sec:inverse_boundary}
\begin{figure}
    \centering
    \subfloat[]{\includegraphics[scale=0.45]{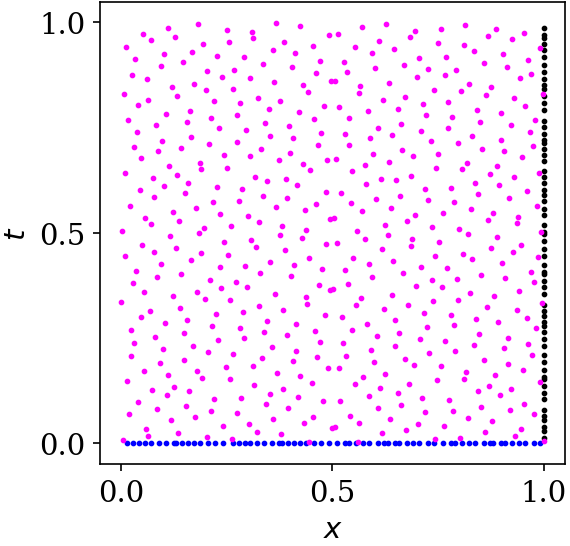}}\hspace{3em}
    \subfloat[]{\includegraphics[scale=0.45]{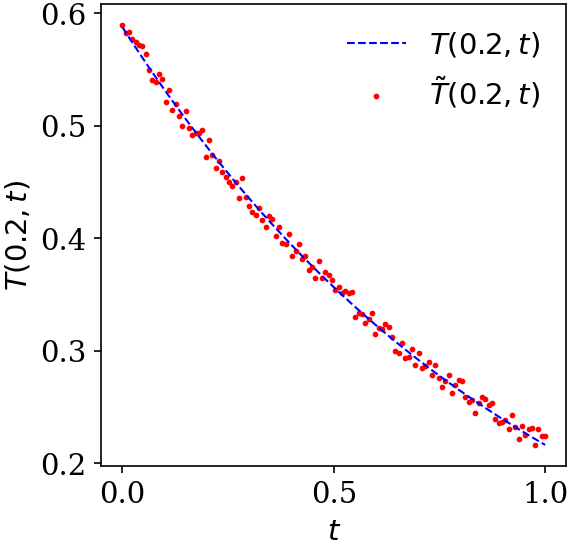}}\hspace{3em}
    \subfloat[]{\includegraphics[scale=0.45]{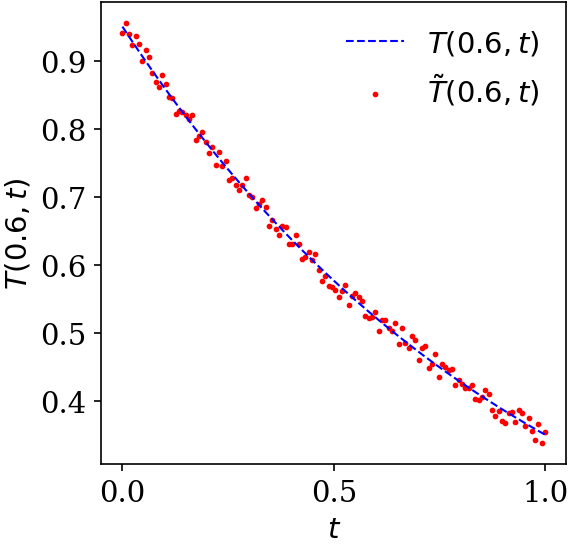}} 
    \caption{Collocation points and measurement data for estimating transient thermal boundary condition from noisy data: (a) collocation points (magenta), boundary points (black) and collocation points for approximating the initial condition (blue), (b) noisy measured temperature $\Tilde{T}$ and exact temperature $T$ at $x=0.2$, (c) temperature $\Tilde{T}$ and exact temperature $T$ at $x=0.6$}
    \label{fig:estimate_thermal_boundayr_data}
\end{figure}

Here we aim to learn the transient thermal boundary condition and temperature based on measured data. Let us consider a parabolic partial differential of the form
\begin{equation}
    \frac{\partial T}{\partial t} = \kappa \frac{\partial^2 T}{\partial x^2} \quad (x,t) \in (0,1)  \times (0,1],
    \label{eq:inverse_heat}
\end{equation}
%%%
where $T$ is the temperature, $\kappa$ is the thermal conductivity. The forward solution of Eq.~\ref{eq:inverse_heat} requires the following boundary information 
\begin{equation}
    \begin{aligned}
        T(0,t) ~ \forall t \in (0,1], \quad
        T(1,t) ~ \forall t \in (0,1], \quad
        T(x,0) ~ \forall x \in (0,1).
    \end{aligned}
\end{equation}
However, in this problem, we do not know the boundary condition at $T(0,t)$. Instead we have noisy data at the interior part of our domain. We constrain the prediction of our model on the governing physics and the noiseless boundary condition that we treat as high-fidelity data.  Fig.~\ref{fig:estimate_thermal_boundayr_data}(a)-(c) present the collocation points distributed over the space-time domain, and synthetically generated noisy data from the exact solution at two locations. 

\begin{figure}
    \subfloat[]{\includegraphics[scale=0.62]{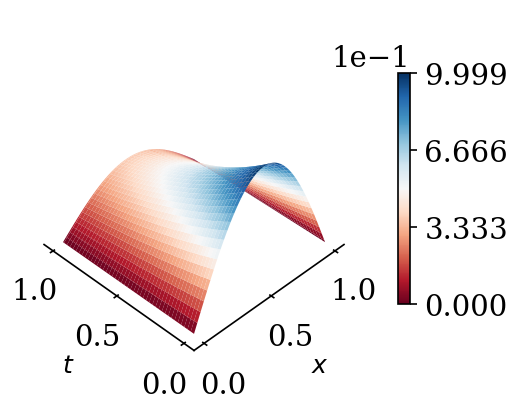}}
    \subfloat[]{\includegraphics[scale=0.62]{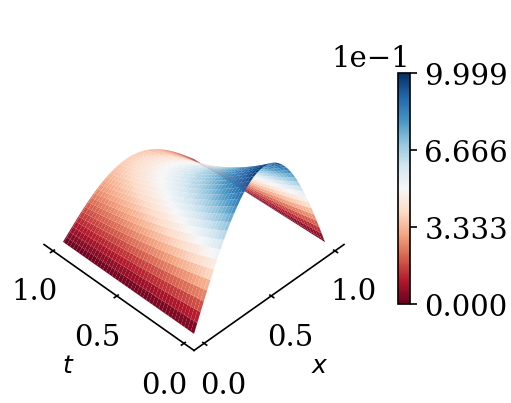}}
    \subfloat[]{\includegraphics[scale=0.62]{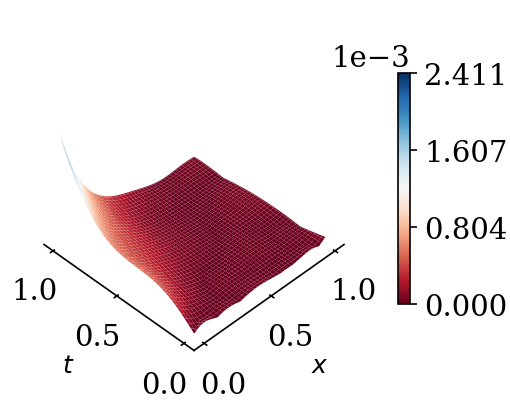}}
    \caption[Estimating transient thermal boundary condition]{Estimating transient thermal boundary condition: (a) exact solution (b), predicted solution (c), absolute pointwise error}
\label{fig:transient_thermal_boundary_estimation}
\end{figure}

We use a fully connected artificial neural networks with three hidden layers and 30 neurons per layer. We adopt the tangent hyperbolic activation function. We randomly generate $N_{\mathcal{F}} = 512$ collocation points, $N_{\mathcal{I}} = 64$ number of training points for approximating the initial condition and $N_{\mathcal{B}} = 64$ number of training boundary data only once before training. For the noisy measurement data we generate $N_{\mathcal{M}} = 2 \times 128$ with  10 percent noise (Gaussian noise with a standard deviation that is 10 $\%$ of the standard deviation of the synthetically generated clean data). 
We use the L-BFGS optimizer with its default parameters and \emph{strong Wolfe} line search activation function. We train our model for $5000$ epochs. The result of our numerical experiment is presented in Fig.~\ref{fig:transient_thermal_boundary_estimation}.

Figure ~\ref{fig:transient_thermal_boundary_estimation}(a) presents the exact solution. From Fig.~\ref{fig:transient_thermal_boundary_estimation}(b) we observe that our predicted solution is in great agreement with the exact solution. We provide the absolute point-wise error distribution in Fig.\ref{fig:transient_thermal_boundary_estimation}(c). We have shown that by having two sensors at two different locations inside the domain that are measuring noisy data, we can sufficiently learn not only the boundary condition, but also the entire temperature distribution across the space-time domain. 

\subsection{Inverse problem: estimation of transient source term in diffusion problems}\label{sec:heat_source_reconstruction}
%%%
In this example, we aim to learn the temperature distribution as well as the heat source using noisy measurements and noiseless boundary or initial conditions, which is a more challenging inverse problem than the previous example. \citet{hasanov2012identification} proposed a collocation algorithm, based on the piece-wise linear approximation of the unknown heat source. Several works have also been proposed to learn the unknown heat source with a single dependent variable \cite{farcas2006boundary,johansson2007determination}. In our approach, we use a neural network model to represent the temperature and the heat source with time and space dependency. We then learn the heat distribution and the heat source simultaneously.

\begin{figure}
    \centering
    \subfloat[]{\includegraphics[scale=0.45]{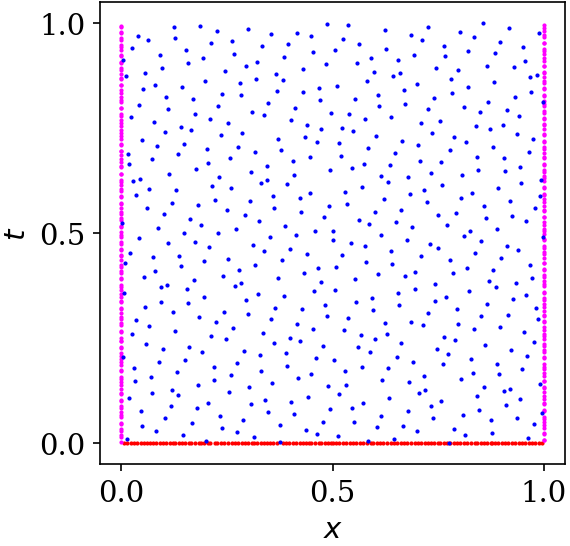}} \hspace{5em}
    \subfloat[]{\includegraphics[scale=0.45]{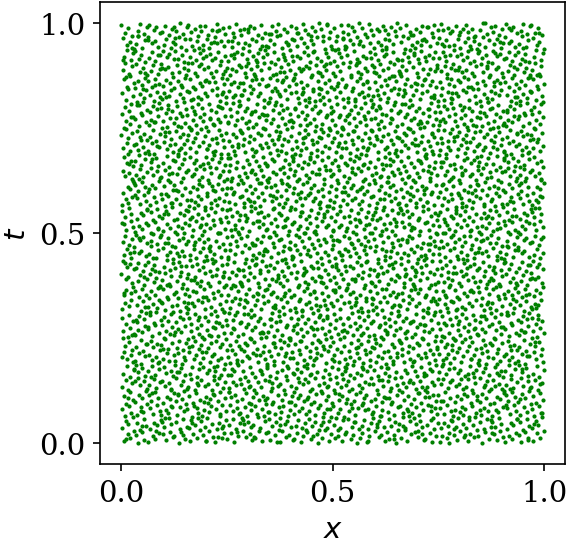}}
    \caption[Positions of training data in space and time]{schematic representation of training data: (a) initial conditions (red points), boundary conditions (magenta points) and noisy measurement data (blue points), (b) collocation points.}  \label{fig:heat_source_reconstruction_data}
\end{figure}

Let us consider the following initial-boundary value problem of the heat conduction equation as follows: 
\begin{equation}
    \frac{\partial u}{\partial t} = \kappa \frac{\partial^2 u}{\partial x^2} + s, \quad (x,t) \in (0,1) \times (0,1],
\end{equation}
where $u$ is the temperature, $\kappa$ is the thermal conductivity and $s$ is the heat source. The objective is to learn the solution $u$ and the heat source $s$ from noiseless boundary condition and noisy measurements inside our domain. A schematic representation of data is given in Fig.~\ref{fig:heat_source_reconstruction_data}.
\begin{figure}
    \centering
    \subfloat[]{\includegraphics[scale=0.63]{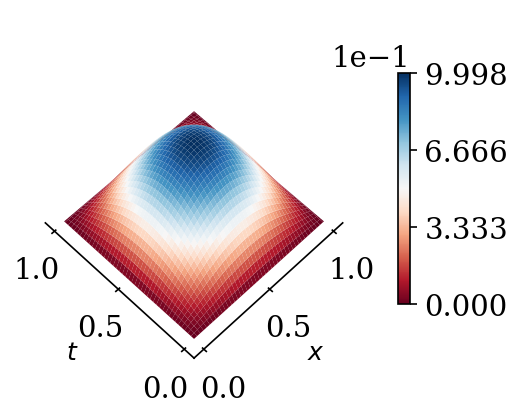}}
    \subfloat[]{\includegraphics[scale=0.63]{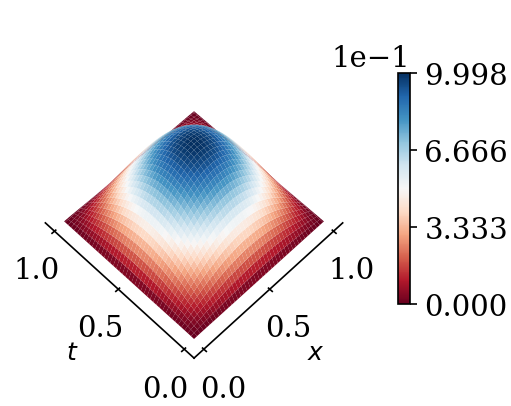}}
    \subfloat[]{\includegraphics[scale=0.63]{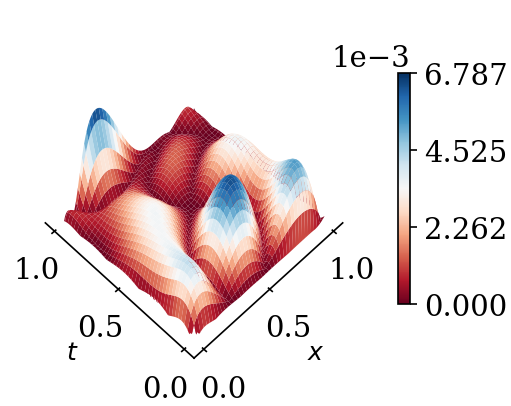}}\\
    
    \subfloat[]{\includegraphics[scale=0.63]{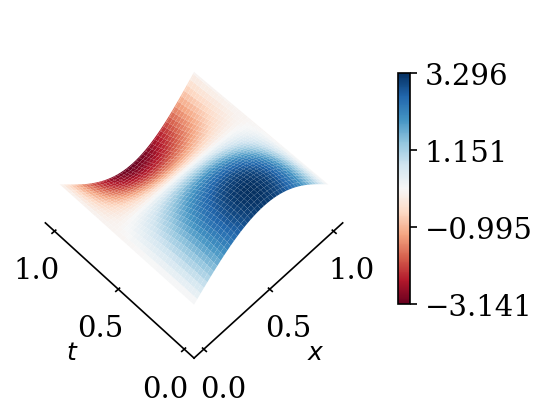}}
    \subfloat[]{\includegraphics[scale=0.63]{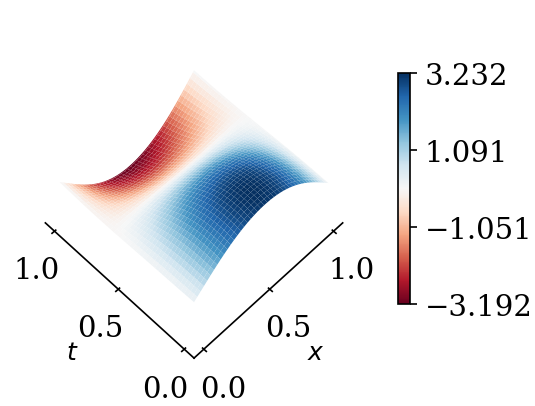}}
    \subfloat[]{\includegraphics[scale=0.63]{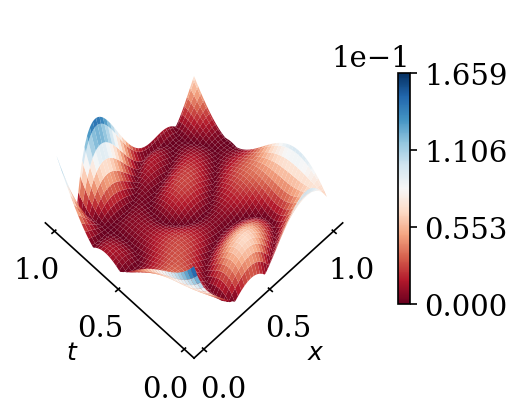}}
    \caption[Estimating transient heat source]{Estimating transient heat source:  (a) exact temperature distribution, (b) predicted temperature distribution, (c) absolute point wise error in temperature distribution, (d) exact heat source distribution, (e) predicted heat source distribution, (f) absolute point wise error in heat source distribution}
    \label{fig:transient_heat_source_estimation}
\end{figure}
Unlike typical inverse source problems that rely on a single input variable, this paper addresses a more challenging scenario where the problem is dependent not only on the spatial variable x but also on the temporal variable t. We demonstrate that we obtain an accurate prediction without resorting to any stabilizing scheme or assuming any fundamental solution for the temperature or the heat source distribution. It is worth mentioning that our constrained optimization problem may also be called as PDE-constrained optimization problem. For this problem, we use three-hidden layer fully connected artificial neural networks with 30 neurons per layer with tangent hyperbolic activation function. We randomly generate $N_\mathcal{F} = 4096$ collocation points, $N_\mathcal{I} = 128$ number of training points for approximating the initial condition and $N_\mathcal{B} = 2\times 128$ number of training boundary data only once before training. For the noisy data we generate $N_{\mathcal{M}} = 512$ with  10 percent noise (Gaussian noise with a standard deviation that is 10 $\%$ of the standard deviation of the synthetically generated clean data). We use L-BFGS optimizer with its default parameters and \emph{strong Wolfe} line search activation function. We train our model for $5000$ epochs. 

We present the result from this numerical experiment is presented in Fig.\ref{fig:transient_heat_source_estimation}. The exact temperature distribution is plotted in Fig.~\ref{fig:transient_heat_source_estimation}(a). From Figs.~\ref{fig:transient_heat_source_estimation}(b)-(c) we observe that our predicted solution is in agreement with the exact solution to a great degree. The exact distribution of the heat source is plotted in Fig.\ref{fig:transient_heat_source_estimation}(d).  Similar to the temperature distribution, the predicted heat source distribution agrees closely with the exact solution as shown in Figs.\ref{fig:transient_heat_source_estimation}(e)-(f).

\section{Conclusion}\label{sec:Conclusion}
Constrained optimization is central to the formulation of the PECANN framework \citep{PECANN_2022} for learning the solution of forward and inverse problems and sets it apart from the popular PINN framework, which adopts an unconstrained optimization in which different loss terms are combined into a composite objective function through a linear scalarization. In the PECANN framework, the augmented Lagrangian method (ALM) enables us to constrain the solution of a PDE with its boundary and initial conditions and with any high-fidelity data that may be available in a principled fashion. However, for challenging forward and inverse PDE problems, the set of constraints can be large and diverse in their characteristics. In the present work, we have demonstrated that conventional approaches to update penalty and Lagrange multipliers within the ALM can fail for PDE problems with complex constraints. Consequently, we introduced an adaptive ALM in which a unique penalty parameter is assigned to each constraint and updated according to a proposed rule. This new method within the PECANN framework enables us to 
constrain the solution of PDEs flexibly toward convergence without the need to manually tune the hyperparameters in the objective function or resort to alternative strategies to balance the loss terms, which may not necessarily satisfy the optimality conditions. 

When the problem size gets larger, constraining each collocation point associated with a constraint becomes taxing computationally. We have empirically demonstrated that Lagrange multipliers associated with a particular type of constraint exhibits a distribution with a clear mean value. Consequently, we have revised our constrained optimization formulation based on the expectation of each loss and constraint terms. Based on our experiments, the resulting formulation does not affect the efficacy of our PECANN framework, but makes it computational more efficient and enable mini-batch training. 

We applied our proposed model to several forward and PDE-constrained inverse problems to demonstrate its efficacy and versatility. A noteworthy example is our simulation of the lid-driven cavity benchmark problem at a Reynolds number 1000 using the primitives-formulation of the Navier-Stokes equations. The $Re=1000$ case is known to be a challenging case to simulate because of the dominant effect of advection in the flow physics and development of strong circulations in the bottom corners of the cavity. Unlike, streamfunction-vorticity formulation of the two-dimensional Navier-Stokes equation, primitive-variables formulation is more general, but at the same more challenging to solve computationally because the formulation retains pressure as one of the variables in addition to the velocity field and. In our flow simulations, we found that the key to obtaining results that are in close agreement with benchmark data \cite{ghia_1982} was to constrain the momentum equations with the divergence-free condition, boundary conditions, and with a small set of randomly sampled high-fidelity data while handling each type constraint with its own unique, adaptive penalty parameter. Another PDE problem that has been viewed as a challenge case for PINNs is the one-dimensional wave equation. Again, our proposed model performed better than the state-of-the-art PINN models for the wave equation problem as well. The codes used to produce the results in this paper are publicly available at \url{https://github.com/HiPerSimLab/PECANN}. Two additional examples are provided in the Appendix.

\section*{Acknowledgments}
This material is based upon work supported by the National Science Foundation under Grant No. 1953204 and in part in part by the University of Pittsburgh Center for Research Computing through the resources provided.
%%%%
\appendix
\section{Additional Examples}\label{sec:appendix}
%%%
\subsection{Inviscid scalar transport in one dimension}
In this section, we study the transport of a physical quantity dissolved or suspended in an inviscid fluid with a constant velocity. Numerical solutions of inviscid flows often exhibit a viscous or diffusive behavior owing to numerical dispersion. Let us consider a convection equation of the form
\begin{align}
        \frac{\partial \xi}{\partial t} + u \frac{\partial \xi}{\partial x} &= 0, ~\forall (x,t) \in \Omega \times [0,1],
        \label{eq:convectionPDE}
\end{align}
satisfying the following boundary condition 
\begin{align}
    \xi(0,t) &= \xi(2 \pi ,t) ~\forall t \in [0,1],
\end{align}
and initial condition
\begin{align}
        \xi(x,0) &= h(x), ~ \forall x \in \partial \Omega,
\end{align}
where $\xi$ is any physical quantity to be convected with the velocity $u$,  $\Omega = \{x~|~ 0 < x < 2\pi \}$ and $\partial \Omega$ is its boundary. Eq.\eqref{eq:convectionPDE} is inviscid, so it lacks viscosity or diffusivity. 
For this problem, we consider $u=40$ and  $h(x) = \sin(x)$. The analytical solution for the above problem is given as follows \cite{krishnapriyan2021characterizing}
\begin{equation}
    \xi(x,t) = F^{-1}(F(h(x) )e^{-i u k t}),
\end{equation}
where $F$ is the Fourier transform, $k$ is the frequency in the Fourier domain and $i = \sqrt{-1}$. Since the PDE is already first-order, there is no need to introduce any auxiliary parameters. Following is the residual form of the PDE used to formulate our objective function,
\begin{subequations}
\begin{align}
    &\mathcal{F}(x,t) = \frac{\partial \xi(x,t)}{\partial t}
    + u \frac{\partial \xi(x,t)}{\partial x}, \\
    &\mathcal{B}(t)= \xi(0,t) - \xi(2\pi,t),\\
    &\mathcal{I}(x) = \xi(x,0) - \sin(x),
\end{align}
\label{eq:residual_form_convection_equation}
\end{subequations}
where $\mathcal{F}$ is the residual form of our differential equation, $\mathcal{B}$ and $\mathcal{I}$ are our boundary condition and initial condition constraints. 

%\citet{krishnapriyan2021characterizing}, proposed curriculum learning as a technique to progressively increase the parameter $u$ until the desired value is reached during training, which proved effective in solving a problem where the complexity of learning the solution $\xi(x,t)$ increased with $u$. However, for a general PDE, such as Poisson's equation, it may not be obvious which parameter to progressively increase in order to improve the learning process. In contrast, our proposed method significantly outperforms curriculum learning, achieving several orders of magnitude improvement in error norms as we will demonstrate later. 

\begin{figure}
    \subfloat[]{\includegraphics[scale=0.55]{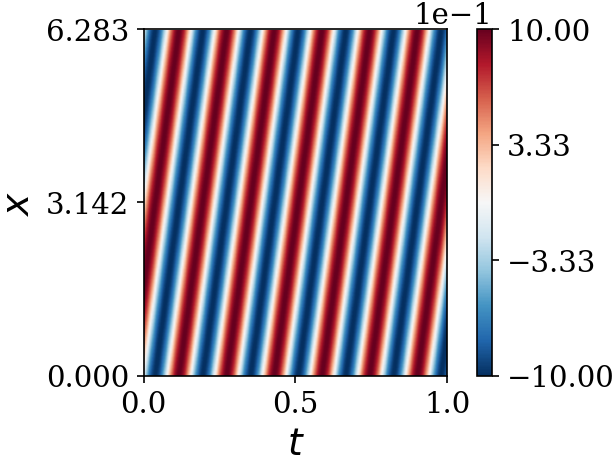}}
    \subfloat[]{\includegraphics[scale=0.55]{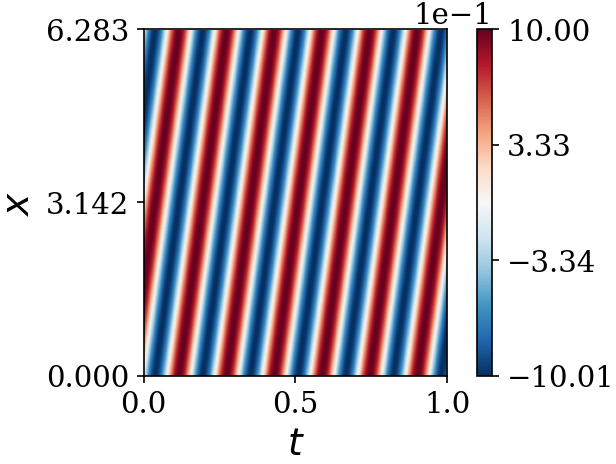}}
    \subfloat[]{\includegraphics[scale=0.55]{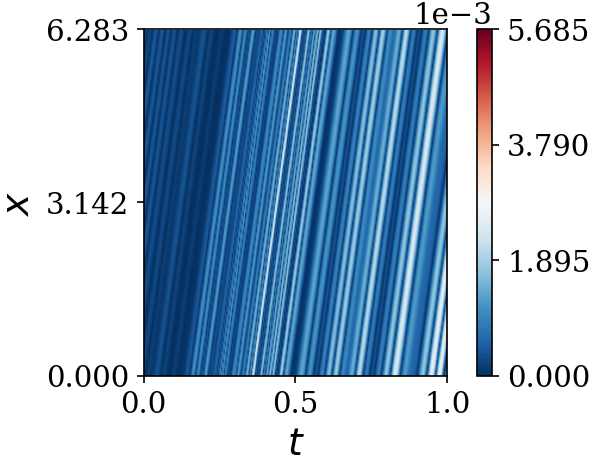}}
    \caption[Convection equation]{Inviscid scalar transport problem: (a) exact solution, (b) predicted solution, (c) absolute point-wise error}
    \label{fig:proposed_convection_equation}
\end{figure}

For this problem, we use a fully connected neural network architecture consisting of four hidden layers with 50 neurons and tangent hyperbolic activation functions. We generate $N_{\mathcal{F}} = 512$ collocation points from the interior part of the domain, $N_{\mathcal{B}} = 512$ from each boundary, and $N_{\mathcal{I}} = 512$ for approximating the initial conditions at each epoch. We use L-BFGS  optimizer with its default parameters and \emph{strong Wolfe} line search function that is available in the PyTorch package. We train our network for 5000 epochs. We present the prediction of our neural network in Fig.\ref{fig:proposed_convection_equation}. We observe that our neural network model has successfully learned the underlying solution as shown in Figs.~\ref{fig:proposed_convection_equation}(b)-(c). 

\begin{table}[b]
\centering
\caption{Inviscid scalar transport problem: summary of the relative error $\mathcal{E}_r$ and the mean absolute error (MAE) metrics for training a fixed neural network architecture.}
\label{tb:Convection}
\vspace{2pt}
\resizebox{0.5\textwidth}{!}{%
\begin{tabular}{@{}lrrrrr@{}}
\toprule
\multicolumn{1}{c}{Models} &
  \multicolumn{1}{c}{$\mathcal{E}_r(\xi,\hat{\xi})$} &
  \multicolumn{1}{c}{MAE} &
  \\ \midrule
Baseline PINN \cite{krishnapriyan2021characterizing}  & $9.61 \times 10^{-1}$& $5.82 \times 10^{-1}$ \\
Curriculum learning \cite{krishnapriyan2021characterizing}  & $5.33 \times 10^{-2}$& $2.69 \times 10^{-2}$ \\
Current method & $\boldsymbol{8.161 \times 10^{-4}}$ & $\boldsymbol{6.810 \times 10^{-4}}$ \\
\end{tabular}}
\end{table}

We also present a summary of the error norms from our approach and state-of-the-art results given in \cite{krishnapriyan2021characterizing} in Table~\ref{tb:Convection}. We observe that our method achieves a relative $\mathcal{E}_r = 8.161 \times 10^{-4}$, which is two orders of magnitude better than curriculum learning method presented in \cite{krishnapriyan2021characterizing}, and three orders of magnitude better than the baseline PINN model performance as reported in the same study.

\subsection{Mixing of a hot front with a cold front}\label{sec:mixing_flow}
\begin{figure}
 \centering
    \subfloat[]{\includegraphics[scale=0.50]{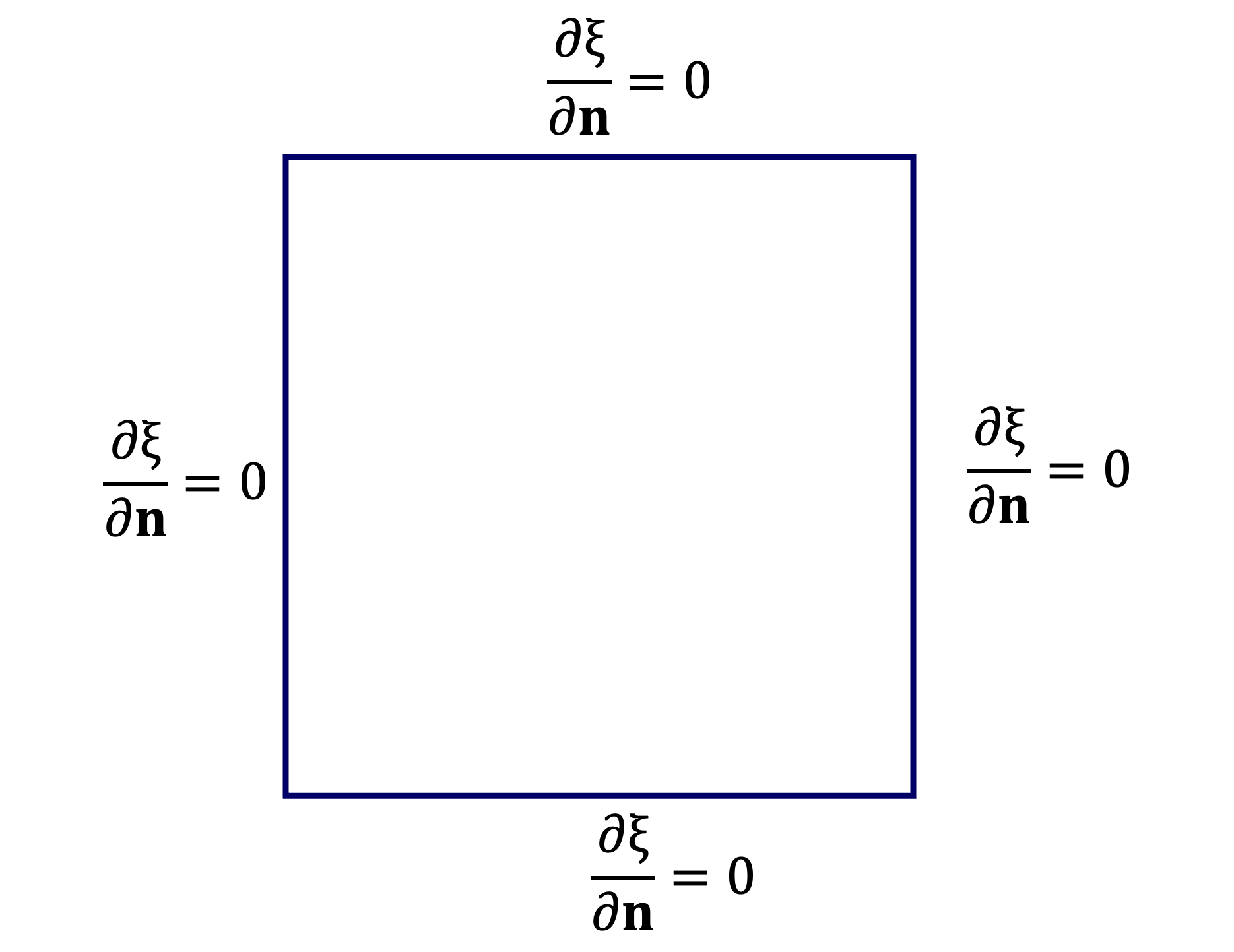}}
    \caption{Mixing hot and cold front: Schematic representation of the domain with its boundary condition}
    \label{fig:domain_mixing_hot_cold}
\end{figure}

\begin{figure}[!h]
 \centering
    \subfloat[]{\includegraphics[scale=0.48]{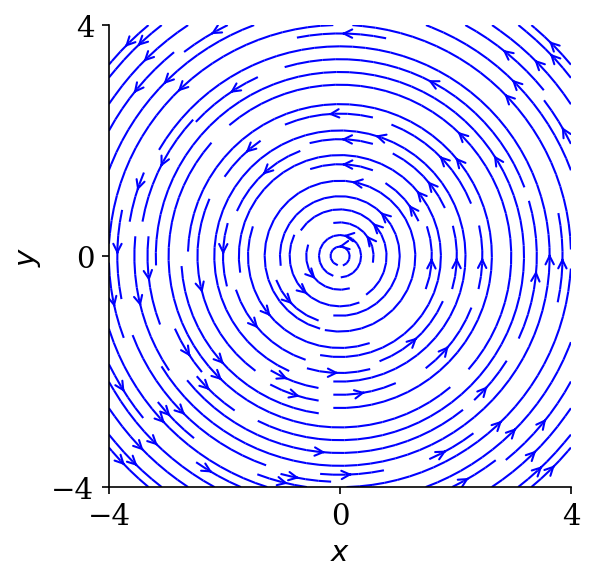}} \hspace{5em}
    \subfloat[]{\includegraphics[scale=0.50]{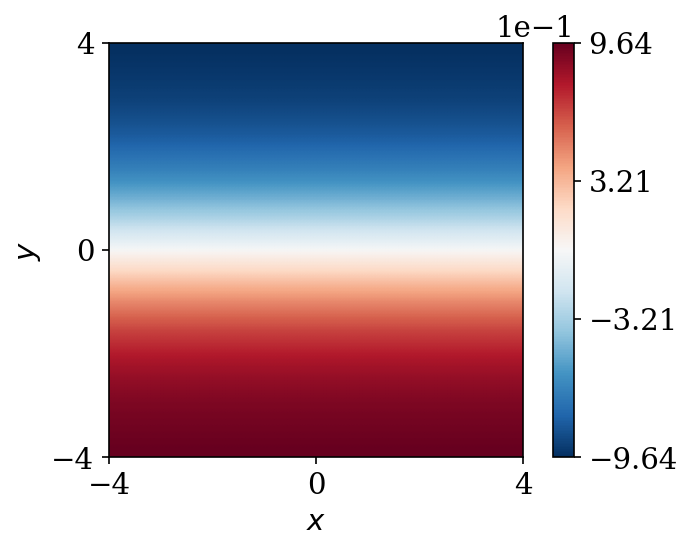}}
    \caption[Velocity and initial state of mixing of hot and cold fronts]{Velocity and initial state of mixing of hot and cold fronts: (a) velocity field, (b) initial state of our flow}
    \label{fig:mixing_flow_field}
\end{figure}

%%%%
In this section, we explore formation of cold and warm fronts in a two-dimensional environment which can be described by the following convection equation: 
\begin{equation}
    \mathcal{F}:=\frac{\partial \xi}{\partial t} + u \frac{\partial \xi }{\partial x} + v \frac{\partial \xi }{\partial y} = 0 ~ \in \Omega \times (0,T],
\end{equation}
where $\Omega = \{(x,y) ~|~ -4 \le x,y \le 4 \}$ and $T=4$. The boundary conditions considered throughout this work are zero flux of the variable $\xi$ along each boundary as shown in Fig.~\ref{fig:domain_mixing_hot_cold}.
The problem has the following analytical solution:
\begin{equation}
    \xi(x,y,t)=-\tanh{\left[ \frac{y}{2}\cos{\omega t} - \frac{x}{2}\sin{\omega t}\right]},
\end{equation}
where $\omega$ is the frequency, which is defined as
\begin{equation}
    \omega=\frac{\nu_t}{r\nu_{t, max}}.
\end{equation}
The velocity field 
$\nu_{t} = \sech{r} ^2 \tanh{r}$
is the tangential velocity around the center with a maximum value ${\nu_{t}}_{\max} = 0.385$.  The velocity components in the $x$ and $y$ directions can be obtained, respectively, as follows: 
\begin{align}
    u(x,y) = -\omega y, \quad v(x,y) =  \omega x.
\end{align} 
We also present the velocity field in Fig.\ref{fig:mixing_flow_field}(a). The initial scalar field varies gradually from positive values at the bottom to negative values at the top as can be seen Fig.\ref{fig:mixing_flow_field}(b). 
where the maximum and minimum values of the field is $\xi_{\max}= 0.964$ and $\xi_{\min}= -0.964$,
respectively.

\begin{figure}[!h]
 \centering
    \subfloat[]{\includegraphics[scale=0.50]{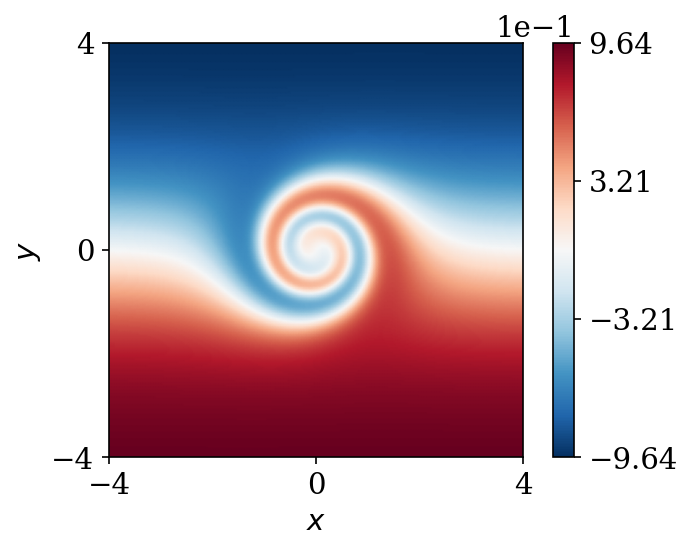}}\hspace{3em}
    \subfloat[]{\includegraphics[scale=0.50]{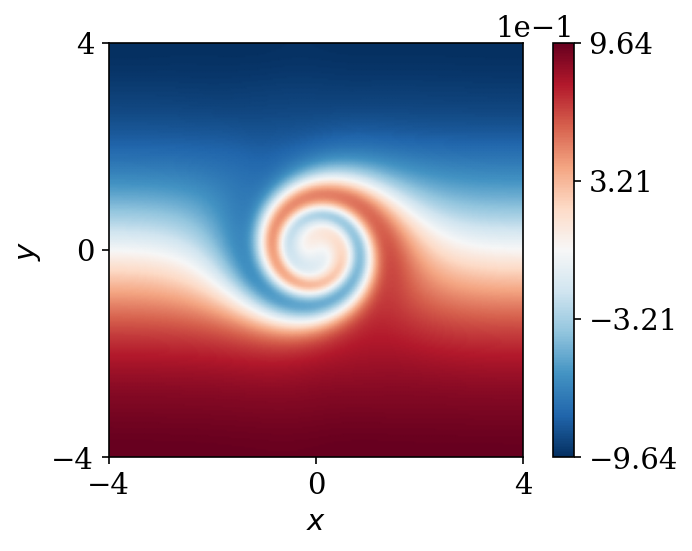}}
    \caption[Mixing of a hot front with a cold front]{Mixing of a hot front with a cold front: (a) exact final state of the scalar field, (b) predicted final state}
    \label{fig:mixing_flow}
\end{figure}
 We use a deep fully connected neural network with 4 hidden layers each with 30 neurons that we train for 5000 epochs total. We use a Sobol sequence to generate $N_\mathcal{F} = 10,000$ residual points from the interior part of the domain, $N_\mathcal{B} = 512$ points from each of the boundaries (faces) and $N_\mathcal{I} = 512$ points for enforcing the initial condition only once before training. Our optimizer is LBFGS with its default parameters and \emph{strong Wolfe} line search function that is available in the PyTorch package.  
 
 Figure \ref{fig:mixing_flow} presents the temperature contours obtained from exact solution along with our predictions. We observe that our PECANN model is in good agreement with the exact solution. We also provide a summary of RMS errors obtained from our neural network model along with conventional numerical methods used in other works in Table.~\ref{tb:Mixing_flow}.

\begin{table}[!h]
\centering
\caption{Mixing of a hot front with a cold front. Comparison of RMS errors.}
\label{tb:Mixing_flow}
\vspace{2pt}
\resizebox{\textwidth}{!}{%
\begin{tabular}{@{}lrrrrrr@{}}
\toprule
\multicolumn{1}{c}{Schemes} &
 \multicolumn{1}{c}{$16 \times 16$} &
\multicolumn{1}{c}{$32 \times 32$} &
\multicolumn{1}{c}{$64 \times 64$} &
\multicolumn{1}{c}{$128 \times 128$} &
\multicolumn{1}{c}{$256 \times 256$} &
 \\ \midrule
First-order upwind (FOU) \citep{tamamidis1993evaluation}& $9.53 \times 10^{-2}$ & $7.89\times 10^{-2}$ & $6.10\times 10^{-2}$ & $4.33\times 10^{-2}$ & $2.18 \times 10^{-2}$ \\
Second-order upwind (SOU) \citep{tamamidis1993evaluation} & $7.92\times 10^{-2}$ & $5.31\times 10^{-2}$ & $2.45\times 10^{-2}$ & $6.4\times 10^{-3}$ &$1.6\times 10^{-3}$ \\
Quadratic upstream-biased (QUICK) \citep{tamamidis1993evaluation} & $6.70\times 10^{-2}$& $3.96\times 10^{-2}$ &$1.23\times 10^{-2}$ &$3.2\times 10^{-3}$ &$8.0\times 10^{-4}$ \\
Van Leer's (MPL) \citep{tamamidis1993evaluation} &$ 7.00\times 10^{-2}$ & $4.70\times 10^{-2}$ &$2.18\times 10^{-2}$ & $1.05\times 10^{-2}$ & $5.4\times 10^{-3}$ \\
Monotonic second-order upwind (MSOU) \citep{tamamidis1993evaluation} & $6.76\times 10^{-2}$ & $4.26 \times 10^{-2}$ & $1.27 \times 10^{-2}$ & $3.9\times 10^{-3}$ & $1.4\times 10^{-3}$ \\
\textbf{Current method} & $\boldsymbol{3.617 \times 10^{-3}}$ & $\boldsymbol{3.550 \times 10^{-3}}$ &$\boldsymbol{3.578 \times 10^{-3}}$ & $\boldsymbol{3.571 \times 10^{-3}}$ &$\boldsymbol{3.573 \times 10^{-3}}$
\end{tabular}}
\end{table}

Table~\ref{tb:Mixing_flow} compares the RMS error level of our predictions against the RMS error levels produced by five different finite-volume based advection schemes as presented in the work of \citet{tamamidis1993evaluation}. Among the finite-volume based methods, the most accurate numerical results were obtained with the QUICK scheme on all meshes consistently. We should note that our results on the Cartesian mesh sizes shown in Table~\ref{tb:Mixing_flow} are post calculations from the trained neural network. For this reason, our error levels are nearly the same across all mesh sizes.

\bibliographystyle{plainnat}
\bibliography{citations}
\end{document}